# MetaDent: Labeling Clinical Images for Vision-Language Models in Dentistry

M.-X. Li[1,2†] 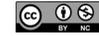, W.-H. Deng[3†] , Z.-X. Wu[1], C.-X. Jin[1], J.-M. Wu[4], Y. Han[5], J. K. H. Tsoi[4] , G.-S. Xia[2*], and C. Huang[1*]

## Abstract

Vision-language models (VLMs) have demonstrated significant potential in medical image analysis, yet their application in intraoral photography remains largely underexplored due to the lack of fine-grained, annotated datasets and comprehensive benchmarks. To address this, we present MetaDent, a comprehensive resource that includes 1) a novel and large-scale dentistry image dataset collected from clinical, public, and web sources; 2) a semistructured annotation framework designed to capture the hierarchical and clinically nuanced nature of dental photography; and 3) comprehensive benchmark suites for evaluating state-of-the-art VLMs on clinical image understanding. Our labeling approach combines a high-level image summary with point-by-point, free-text descriptions of abnormalities. This method enables rich, scalable, and task-agnostic representations. We curated 60,669 dental images from diverse sources and annotated a representative subset of 2,588 images using this meta-labeling scheme. Leveraging large language models (LLMs), we derive standardized benchmarks: approximately 15,000 visual question answering (VQA) pairs and an 18-class multilabel classification dataset, which we validated with human review and error analysis to justify that the LLM-driven transition reliably preserves fidelity and semantic accuracy. We then evaluate state-of-the-art VLMs across VQA, classification, and image captioning tasks. Quantitative results reveal that even the most advanced models struggle with a fine-grained understanding of intraoral scenes, achieving moderate accuracy (e.g., less than 70% in VQA) and producing inconsistent or incomplete descriptions in image captioning. These findings underscore the gap between general-purpose VLMs and the demands of specialized models, highlighting the need for domain-adapted training and more sophisticated evaluation protocols to assist professional dental practice and community oral health efforts. We publicly release our dataset, annotations, and tools to foster reproducible research and accelerate the development of vision-language systems for dental applications.

## Introduction

Dental photography plays a crucial role in diagnosis, treatment planning, patient education, and documentation across dental disciplines (Ding et al 2023; Caron et al 2025; Mania et al 2025). In this context, recent advancements in vision-language models (VLMs) offer a compelling opportunity for automated and scalable image interpretation. By leveraging large-scale annotated datasets, VLMs have demonstrated remarkable capabilities in understanding and reasoning about visual content in natural images (Radford et al 2021; Liu et al 2023).

In health care research, VLMs have been increasingly utilized in analyzing visual and textual data, aiding professionals in detection, diagnosis, and reporting to improve efficiency (Li et al 2023; Ghosh et al 2024; Ryu et al 2025). Despite these advances, few studies report VLM performance for dental image understanding. A concurrent study established a benchmark for panoramic radiograph analysis and found no significant difference between general-purpose and medical-specific VLMs in dentistry (Hao et al 2025). Another study collected a mixture of large-scale X-ray and intraoral images and fine-tuned a VLM using derived data mainly from categorical

[1]Department of Prosthodontics, State Key Laboratory of Oral & Maxillofacial Reconstruction and Regeneration, Key Laboratory of Oral Biomedicine, Ministry of Education, Hubei Key Laboratory of Stomatology, School and Hospital of Stomatology, Wuhan University, Wuhan, Hubei, China
[2]School of Artificial Intelligence, Wuhan University, Wuhan, Hubei, China
[3]School of Computer Science, Wuhan University, Wuhan, Hubei, China
[4]Dental Materials Science, Clinical Artificial Intelligence, Faculty of Dentistry, The University of Hong Kong, Hong Kong SAR, China
[5]School of Electrical and Computer Engineering, Purdue University, West Lafayette, IN, USA

[†]Authors contributing equally as co-first authors.
[*]Authors contributing equally as co-senior authors.

A supplemental appendix to this article is available online.

**Corresponding Authors:**
C. Huang, School & Hospital of Stomatology, Wuhan University, 237 Luoyu Rd., Hongshan District, Wuhan, Hubei 430079, China.
Email: huangcui@whu.edu.cn
G.-S. Xia, School of Artificial Intelligence, Wuhan University, 299 Bayi Rd., Wuchang District, Wuhan, Hubei 430072, China.
Email: guisong.xia@whu.edu.cn

labeling (Meng et al 2025). While the result looks promising, the dataset and the model are not publicly available.

Based on our preliminary observations, even state-of-the-art VLMs struggle with a fine-grained understanding of intraoral images. We attribute this limitation to 2 interrelated challenges:

1. Dental diagnosis often requires nuanced, hierarchical interpretations that go beyond closed-set classification. Current VLMs, trained primarily on categorical labels or vague text descriptions, lack the capacity to reason over such fine-grained and clinically contextual features.
2. The development and evaluation of VLMs for dentistry are severely constrained by the scarcity of well-annotated, diverse, and publicly available datasets. Most existing datasets for intraoral images focus on specific categories (Dot et al 2024; Nguyen et al 2025; Wang et al 2025) and lack diversity across different data sources (Huang et al 2024; Uribe et al 2024).

To address the above limitations, we curated a dedicated dataset and benchmarked state-of-the-art VLMs, introducing innovations in 2 key aspects: annotation strategy and data sourcing. First, we proposed a novel semistructured labeling strategy that formulates the annotation as an open-set weak labeling task for abnormality detection. This approach provides a compact yet comprehensive representation of each image, which can be reliably translated into various downstream task-specific formats. Second, we constructed a large-scale image dataset with substantial diversity by filtering and curating images from web-scraped data, and then we labeled a subset with the proposed semistructured scheme to enable both precise annotation of abnormalities and holistic scene understanding.

In this work, we introduce MetaDent, a semistructured annotation framework and large-scale resource for vision-language understanding of intraoral images. MetaDent is designed to support diverse downstream tasks—including visual question answering, multilabel classification, and image captioning. The name "MetaDent" reflects its meta-annotation structure, broad data diversity, and adaptability across a wide range of dental artificial intelligence (AI) applications.

To encourage reproducibility and further research, we publicly release the dataset, annotation interface, and benchmarking tools at https://menxli.github.io/metadent/.

## Materials and Methods

### Ethics Statement

The study protocol was approved by the institutional ethics committee of the Hospital of Stomatology, Wuhan University (No. WDKQ2025[C02]). The study adhered to the ethical considerations outlined by the committee, including participant privacy and data protection measures.

### Meta Dataset Curation

We collected images from 3 sources: in-house collected, public dataset, and web-crawled. Specifically, we collected 4,373 clinical photographs from the Department of Prosthodontics at the School of Stomatology, Wuhan University (Data Source 1, DS1). For the public component, we included 9,390 images from the Teeth or Dental image dataset (Data Source 2, DS2) (Chaudhary et al 2024), given its relatively large quantity and good quality. Lastly, we filtered the COYO-700M (Byeon et al 2022), a large-scale web-crawl image corpus, using a fine-tuned ViT-L/16 binary classifier to retain only dental images and an image hasher (Haviana and Kurniadi 2016) for duplicate removal (details in Appendix Section 1). This step resulted in 46,906 images (Data Source 3, DS3). In total, the collected dataset comprises 60,669 dental images.

From this combined image dataset, we randomly sampled a subset of 3,576 images for human review. We excluded images lacking clinical relevance (e.g., artificial images) or sufficient quality (e.g., blurry images), resulting in a final set of 2,588 images for annotation. The annotation follows a semistructured format, where each image is assigned an overall descriptive summary and a list of identified abnormalities (Fig. 1). The overall description is a concise paragraph that introduces the main content of the image as well as the shooting perspective. Abnormalities were defined as any clinically relevant deviations from normal dental anatomy or healthy tissue appearance. Annotators were instructed to list the abnormalities point-by-point using natural, unstructured language focused on visual appearance, including diagnostic interpretations where applicable. Meanwhile, the annotators also draw a rough contour for each abnormality in the image for its corresponding entry.

All images were annotated by 2 dentists with 7 and 4 y of clinical experience, respectively. Prior to formal annotation, the annotators underwent a 1-mo training, during which they received guidance from a senior dentist with 10 y of clinical experience. Training sessions included experimental annotation and corrective feedback. During the formal annotation, any ambiguous or challenging cases were discussed among all 3 experts. When the visual evidence was inconclusive but the annotators had moderate confidence, descriptions were used with cautious wording. When confidence was very low, the entry was flagged as "uncertain." To assess the consistency between the raters, interrater reliability was evaluated. Cohen's $\kappa$ coefficient, calculated on 100 images randomly drawn from the dataset, was 0.83 (Appendix Section 10), indicating a high level of agreement. To facilitate labeling all abnormalities exhaustively, after the initial annotation phase, the 2 annotators cross-verified each other's labels, ensuring that every image was reviewed by at least 2 raters. The initial annotations were in Chinese and translated into English via large language models (LLMs); all LLM-assisted steps in the study were carried out using GPT-OSS-120B (OpenAI et al 2025). Of the 2,588 images processed, the labeling effort yielded 6,314 confidently annotated records and 138 entries flagged as uncertain.

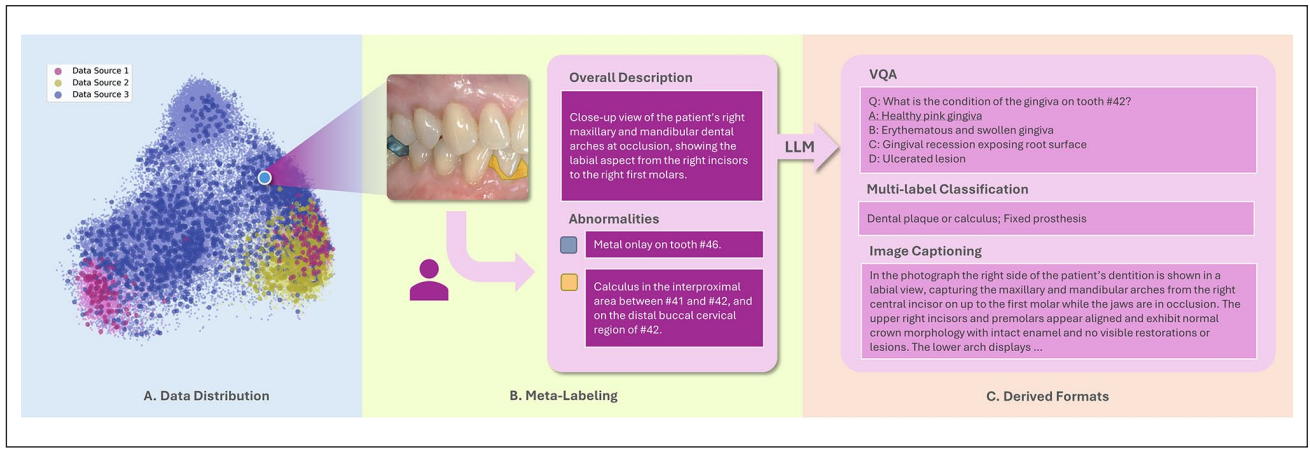

**Figure 1.** Data processing pipeline. (**A**) Composition and distribution of the dataset across 3 sources, with the Internet-scraped subset (Data Source 3) being the largest. Image features were extracted using DINOv3 (Siméoni et al 2025) and projected into 2D using principal component analysis; darker, larger dots represent labeled samples, while lighter and smaller dots indicate the remaining images in the collection. (**B**) The images are compactly labeled in a semistructured pattern as an overall description of the main visual content, as well as point-by-point, free-text descriptions of abnormalities. (**C**) Meta-labels are designed for scalability and can be flexibly converted into various output formats to support downstream tasks by leveraging the reasoning capability of large language models.

## Secondary Dataset Generation

Following the meta-dataset labeling, we converted the semistructured data into standardized formats with the help of an LLM. In particular, we chose 3 tasks: visual question answering (VQA), multilabel classification, and image captioning. This process effectively expands its size and applicability for diverse downstream tasks. The integration of LLM is based on 2 motivations: 1) our semistructured label is, in essence, a compact and complete representation of the intraoral image, which makes the deduction of unmentioned normal structures possible, and 2) current VLMs still face limitations in reliability, particularly in specialized domains (Jeong et al 2024; Nath et al 2025). In contrast, LLM is far more reliable in terms of reasoning and less hallucination when given an appropriate context (Li et al 2024). We harnessed linguistic reasoning to achieve more robust and scalable data processing.

For the VQA generation, we allowed the model to reason about nonmentioned common abnormalities and questioning beyond explicit labels. We generated 5 to 10 questions per image depending on the number of labeled entries. When fewer than 2 abnormalities were labeled in an image, we generated 5 questions; otherwise, 10 would be raised. The question types included judgment (true/false) and multiple-choice. To enhance the quality of the generated VQA pairs, a self-refinement step was applied (Madaan et al 2023). Entries marked as "uncertain" were handled with exclusion from downstream analyses to minimize ambiguity (Appendix Section 5).

For the classification, we established 18 classes based on visual appearance, with clinical pathology serving as a guiding reference (Fig. 4C and Appendix Table 1). For example, "chalky patches" was grouped into "tooth color abnormality," regardless of whether they stemmed from early caries, enamel hypoplasia, fluorosis, or postorthodontic lesions. Likewise, plaque and calculus were sometimes visually indistinguishable in photographs and merged into 1 class. This image-centric approach balances practicality and medical relevance and is presumably more effective than a strict clinical diagnosis from the perspective of image analysis.

For the image captioning, the LLM was prompted to generate free-form descriptions from the meta-labels, which were used as reference captions.

To ensure data quality and to better understand the sources of errors, we analyzed errors arising during the transition, focusing on VQA and multilabel classification. During dataset generation, the LLM was prompted to provide explanations for its answers to support human review. Two annotators evaluated the entire classification dataset and randomly sampled 2 VQA pairs per image to analyze errors and make necessary revisions or deletions. We identified and defined 6 common error types; detailed descriptions of each are provided in Appendix Table 2.

The above steps resulted in 18,416 VQA pairs and 2,588 multilabel classification questions for subsequent assessment. Prompts for generation are supplied in Appendix Section 5.

## Dataset Summary

Figure 2 characterizes MetaDent's visual properties. Most images span $10^5$ to $10^6$ pixels (width: ~200–1,000 px), with aspect ratios varying by source: DS3 shows the broadest spread, with DS2 the most uniform (Fig. 2A, B). Analysis by Ovis (Lu et al 2024) reveals 80% depict humans; among these, eyes and nose are detected for fine-grained categorization (Fig. 2C). Intraoral views comprise 63.5% of the dataset and facial images 5.0% (deidentified upon release). In the labeled subset, most images contain 1 to 3 annotated abnormalities (Fig. 2D). In the Table, we compare MetaDent against existing intraoral image datasets. The key gap that MetaDent uniquely fills is not merely dataset size but the combination of large-scale coverage, semistructured meta-annotation, and broad task generality within a single unified framework.

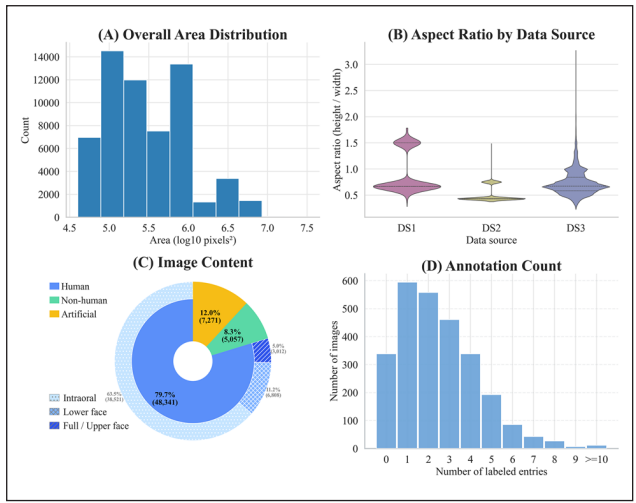

**Figure 2.** Dataset statistics. (**A**) Overall image area distribution: Image size (in log10 pixels²) shows that most images fall within the 5.0 to 6.0 range. (**B**) Aspect ratio by data source: DS3 exhibits the widest distribution of aspect ratios, followed by DS1, whereas DS2 displays a more uniform aspect ratio profile. (**C**) Image content composition: Approximately 80% of images contain human subjects. Of the entire dataset, 63.5% are presumed to be intraoral photographs, and 5.0% show human faces that underwent deidentification prior to dataset release. The presence of key facial landmarks (eyes and nose) is used to further categorize the human-subject images. (**D**) Annotation count of the labeled subset: Most images contain 1 to 3 labeled abnormalities, as shown by the distribution of annotations per image.

### Evaluation of the VLMs

We evaluate VQA performance using accuracy, defined as the proportion of questions answered correctly. This evaluation is performed separately for multiple-choice and true/false question types. For multilabel classification, we report precision, recall, and F1-score, along with Exact Match—the predicted labels exactly match the ground truth. The F1-score, defined as the harmonic mean of precision and recall, reflects the balance between false positives and false negatives. In image captioning, we evaluate the generated captions using LLM-as-a-judge from semantic and diagnostic consistency perspectives. Semantically, we prompt the LLM to generate a reference caption and compare it with the outputs from the VLMs using BERTScore (Zhang et al 2020), which computes token-level contextual embeddings and measures how well the generated caption semantically aligns with the reference description. At the diagnostic consistency level, we instructed the LLM to evaluate whether the generated caption correctly identifies the key diagnostic findings and anatomical location of abnormalities, as specified in the meta-label. This essentially transforms the evaluation into an open-set multilabel classification task. Metrics were computed both per data source and over the full dataset. Detailed definitions of each metric are provided in Appendix Section 2.

The VLMs included in our study represent the current frontiers in vision-language understanding, including proprietary models: GPT-4o-2024-08-06 (OpenAI et al 2024) and Gemini-2.5-flash (Comanici et al 2025), as well as leading open-source models: Qwen3-VL-235B-A22B-Thinking (Yang et al 2025), Ovis2-34B (Lu et al 2024), and Baichuan-Omni-1.5 (Li et al 2025). Several models occasionally produced nonresponses or format-inconsistent outputs; these cases were excluded from analysis, and the exclusion rate is reported in Appendix Table 13. To ensure VLMs did not benefit systematically from sharing linguistic priors with the LLM used in benchmark construction, we repeated the evaluation using an alternative LLM—results showed no noticeable bias (Appendix Section 9).

## Results

### Datasets

In Figure 1A, image features were extracted using DINOv3 (Siméoni et al 2025) and subsequently dimensionally reduced using principal component analysis (PCA). In the plot, labeled data points are represented by darker, larger dots, while unlabeled raw images appear as lighter, smaller dots. The visualization reveals that the Internet-scraped images (DS3) dominate the dataset and occupy a broader region, indicating they introduce greater visual diversity and cover a wider range of scenarios. Moreover, the labeled data were sampled uniformly from the entire distribution. This justifies that the annotated subset is representative of the overall dataset without significant sampling bias.

To assess benchmark quality, we conducted a human review of the VQA subset (5,176 randomly sampled question-answer pairs from 18,416 total) and the entire classification dataset (2,588). Annotators rated 94.2% (4,875) of VQA pairs as correct; for classification, 90.5% of images (2,340) had perfectly correct labels, while 248 (9.5%) contained 1 or more errors. This evaluation confirms the high overall fidelity of the LLM-generated datasets from the meta-labels. The error analysis for VQA and classification dataset generation is illustrated in Appendix Figure 4; the incorrect responses spanned a mix of the error types.

Figure 4B shows the manual classification of the randomly sampled labeled subset, reflecting overall pathology prevalence: "malocclusion or dental malalignment" is most common, while "residual root" and "oral ulcer" are less frequent.

### VQA

All evaluated VLMs achieve moderate accuracy in VQA (Fig. 3A, B). True/false questions (TFQ) consistently receive higher accuracy than multiple-choice questions (MCQ). Gemini-2.5-Flash led performance (MCQ 64.1%, TFQ 67.9%), slightly ahead of GPT-4o (60.6% MCQ, 67.4% TFQ). Open-source models lagged behind: Ovis-2 (61.7% MCQ, 67.3% TFQ), Qwen3-VL (60.4% MCQ, 63.9% TFQ), and Baichuan-Omni (57.5% MCQ, 64.5% TFQ). Performance varied by image source: all models performed slightly better on DS3. Nevertheless, results remained suboptimal, as no model surpassed 68% accuracy on any question type, highlighting the difficulty of fine-grained intraoral VQA and the modest advantage of proprietary over open-source models.

**Table.** Comparison between MetaDent and Other Representative Publicly Available Dental Image Datasets, Including Publication Year, Size, Annotation Type, Image Source, Scope, and Whether Each Dataset Supports Classification (CLS), Image Captioning (CAP), and Segmentation (SEG) Tasks.

| Dataset | Year | Size | Annotation Type | Image Source | Scope | CLS | CAP | SEG |
|---|---|---|---|---|---|---|---|---|
| Oral Images Dataset (Nanditha et al 2020) | 2020 | 323 | Image-level category (benign/malignant) | Multicenter | Benign/malignant lesions | √ | | |
| Oral Cancer (Lips and Tongue) (Shivam and Prakrut 2020) | 2020 | 131 | Image-level category (cancer/ noncancer) | Multicenter | Lip/tongue cancer | √ | | |
| Caries-Spectra (Himel et al 2023) | 2023 | 2,000 | Image-level category (adv/early/no caries) | Single-center | Caries | √ | | |
| Teeth or Dental image dataset (Chaudhary et al 2024) | 2024 | 9,562 | Image-level category (8 healthy-tooth views) | Single-center | Image view classification | √ | | |
| AlphaDent (Sosnin et al 2025) | 2025 | 1,320 | Instance segmentation masks (9 pathology classes) | Single-center | Tooth pathology | √ | | √ |
| Annotated intraoral image dataset for dental caries detection (Faizan Ahmed et al 2025) | 2025 | 6,313 | Image-level category (5 healthy-tooth views) | Single-center | Caries detection | √ | | √ |
| CODE—Comprehensive Oral mucosa Database with Explanations (Madan Kumar et al 2025) | 2025 | 4,300 | Image-level category (eight standard sites) | Multicenter | Mucosa disease | √ | √ | √ |
| MetaDent (ours) | 2025 | 60,669 (labeled 2,588) | Semistructured | Single-center and web | General abnormalities | √ | √ | √ |

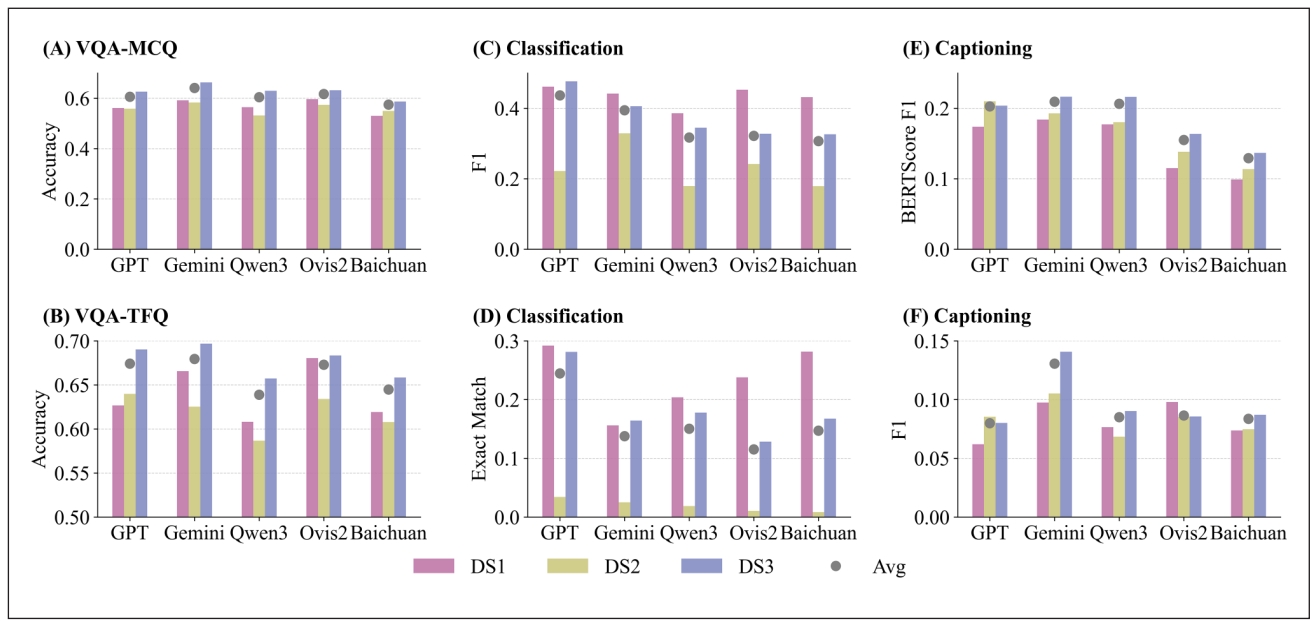

**Figure 3.** Performance of 5 vision-language models on MetaDent across 3 tasks. (**A**) Visual question answering (VQA) accuracy on multiple-choice questions (MCQ). (**B**) VQA accuracy on true/false questions (TFQ). (**C**) F1 for the 18-class multilabel classification task. (**D**) Exact Match accuracy for multilabel predictions. (**E**) Image captioning semantic similarity reported as BERTScore-F1 against reference captions. (**F**) Content accuracy of caption-derived abnormalities, evaluated as multilabel F1 after large language model extraction.

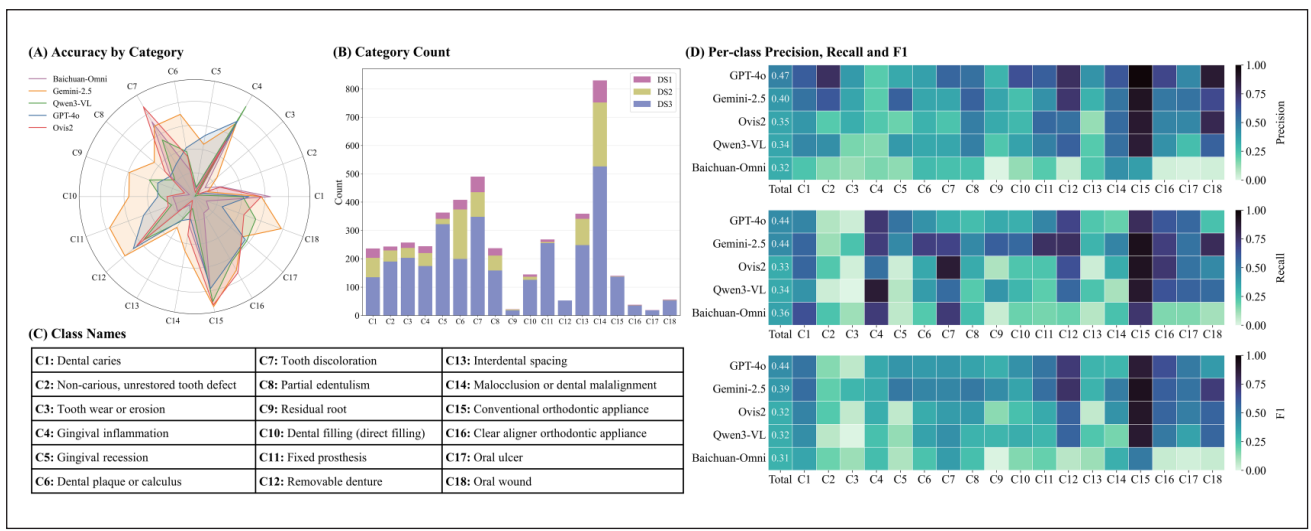

**Figure 4.** Category-level performance for the multilabel classification task. (**A**) Exact Match accuracy across 18 categories for 5 models. (**B**) Distribution of samples per category for the classification task. (**C**) Names of the 18 categories (see Appendix Table 1 for details). (**D**) Per-class precision, recall, and F1 across datasets. The "Total" column reports the overall metrics aggregated across all classes, computed by first evaluating the metric at the sample level and then averaging across all samples.

### Classification

Multilabel classification was challenging for all models (Figs. 3C, D and 4). GPT-4o achieved the highest F1-score (0.437) with balanced precision and recall (0.475/0.438), outperforming Gemini-2.5 (F1 ≈ 0.394, higher recall 0.444 but lower precision 0.398). Open-source models scored lower (F1 ≈ 0.30–0.33). Exact Match accuracy was very low (GPT-4o 24.5%; others <16%), showing models rarely predicted all labels correctly. Overall, even the best model detected less than half of all findings.

### Image Captioning

Models struggled with accurate free-form captions (Fig. 3E, F). Gemini-2.5 achieved the highest semantic similarity to references (BERTScore-F1 ≈ 0.209), followed by Qwen3-VL (0.206), GPT-4o (0.203), Ovis-2 (0.155), and Baichuan-Omni (0.129). Appendix Section 9 reveals slightly different styles: GPT-4o (P ≈ 0.112, R ≈ 0.072) and Ovis (P ≈ 0.108, R ≈ 0.084) favored precise captions but led to less sensitivity (low recall), while the others were more balanced. Consistency remained low across all models (F1 ≈ 13% at best), indicating most key findings were missed or incorrectly interpreted. In summary, captions generated by VLMs are often unreliable for clinical interpretation.

## Discussion

In this study, we focus on the limitations of VLMs for intraoral image understanding and propose a semistructured meta-labeling technique to support analysis of clinical images. Unlike conventional categorical or textual labels, our approach generates rich, hierarchical representations that capture semantic meaning at arbitrary levels of detail, enabling effective transfer to downstream tasks. To support this work, we collected a diverse dataset of dental images primarily from online sources and carefully annotated a subset to evaluate representative VLMs. Our results show that state-of-the-art models struggled across all tasks—barely reaching about 65% to 70% accuracy in VQA and around 0.4 F1 in multilabel classification—and their image captions often miss important findings. We further observe a sharp performance degradation on DS2, suggesting the presence of a domain shift. A more detailed analysis of the challenges associated with DS2 indicates that the performance drop may be attributed to shifts in image tone and demographic differences between datasets (detailed analysis in Appendix Section 3). These findings highlight that current VLMs, even cutting-edge systems such as GPT-4o, are not yet reliable for deployment in clinical settings without further refinement.

In contrast to prior studies that typically rely on photographs from a single institution or a limited number of clinical centers, we adopted a different approach by primarily utilizing web-scraped images. Specifically, we used a filtered subset of the COYO-700M dataset, which was originally sourced from Common Crawl (2008)—a large-scale web archive. This set of images contains diverse photos compared to single-center or public single-source datasets, as shown in Figure 1A. This diversity in imaging conditions, lighting, and patient demographics enhances the generalizability of the dataset, reducing bias and enabling better real-world applicability.

Another contribution of this study lies in the labeling protocol. The advantages of the proposed technique are as follows:

1. It effectively captures the hierarchical complexity of intraoral conditions that conventional categorical labels cannot represent. For example, a full crown may vary in material, color, and defects, which cannot be documented comprehensively with a single category and can be described simultaneously within our structure (Appendix Section 7). This enables richer, more precise supervision signals for VLMs.
2. Compared with free-form captions commonly used in general-domain VLMs, the proposed scheme offers a more efficient and clinically meaningful annotation process. By focusing on abnormalities rather than exhaustive descriptions, it reduces redundancy. Normal conditions can be inferred from the absence of abnormalities, minimizing annotation workload without compromising interpretability. Moreover, the point-by-point label structure also facilitates the attachment of metadata (e.g., bounding boxes, segmentations) for downstream applications.

By balancing comprehensiveness with conciseness, the proposed method provides a compact yet comprehensive annotation of the image by a brief summary of the main visual content and a structured list of free-form abnormalities. By leveraging the strong reasoning capabilities and dental knowledge of LLMs, we can easily scale the dataset to different formats for downstream tasks. These models can not only assist in diagnosis to reduce errors but also enable applications such as smart health care management, early-stage oral disease screening, and self-conducted oral health checks at home. Some of the potential use cases are listed in Appendix Section 8.

The primary limitation of this work lies in the relatively small size of the labeled dataset, which is insufficient to claim broad coverage of intraoral conditions and is hard to support fine-tuning of large, domain-specific VLMs. Additionally, most of the images are web-scraped, introducing variability in quality and uncertain provenance. The labels combine human annotations with LLM-generated outputs, which may introduce noise. That said, LLM-assisted data generation and quality control are common in related research, and we quantitatively assessed error sources to support a certain level of confidence in the dataset's overall reliability. As large models continue to advance, these techniques are expected to yield even higher-quality data. Lastly, while our benchmark is in large quantity and reasonable quality, we acknowledge the absence of direct comparison to human expert performance.

Despite these constraints, our benchmark offers a robust and challenging evaluation platform for multimodal models in dentistry. We release the dataset, an initial set of annotations, together with the labeling tools, hoping this work will encourage community-driven expansion, supporting fine-tuning, regulatory validation, and ultimately aid the development of more robust and practically useful models in oral health.

## Conclusion

In this work, we proposed a semistructured annotation framework for intraoral image analysis that enables rich, scalable, and task-agnostic representations. By curating a diverse dataset and standardized benchmarks, we evaluated state-of-the-art vision-language models and revealed their limitations in fine-grained dental understanding. The result reveals that the gap between current VLM capabilities and dental requirements remains wide. Closing this gap requires concerted efforts and cross-disciplinary collaboration between AI researchers and dental practitioners. By providing the community with a road-map and tools, we aim to help such interdisciplinary synergy to advance multimodal AI in dentistry—ultimately moving us closer to practically useful multimodal AI in oral health care.

## Author Contributions

M.-X. Li, W.-H. Deng, contributed to conception and design, data acquisition, analysis, and interpretation, drafted and critically revised the manuscript; Z.-X. Wu, C.-X. Jin, contributed to data acquisition, analysis, and interpretation, critically revised the manuscript; J.-M. Wu, J. K. H. Tsoi, contributed to data conception and design, critically revised the manuscript; Y. Han, contributed to data analysis, drafted and critically revised the manuscript; G.-S. Xia, C. Huang, contributed to data conception and design, drafted and critically revised the manuscript. All authors gave final approval and agree to be accountable for all aspects of the work.

## Acknowledgments

The authors gratefully acknowledge Chao Pang for helpful discussions and Zhong-Shi Zhang, Yu-Jie Wu, and Mu-Qi Jiang for their valuable contributions to data processing and their unwavering support throughout this work.

## Declaration of Conflicting Interests

The authors declared no potential conflicts of interest with respect to the research, authorship, and/or publication of this article.

## Funding

The authors disclosed receipt of the following financial support for the research, authorship, and/or publication of this article: This study was funded by the National Natural Science Foundation of China (82401200), Natural Science Foundation of Hubei Province (2024AFB033), Key R&D Program of Hubei Provincial Department of Science and Technology (2023BAB058), and National College Students Innovation and Entrepreneurship Training Program (202510486171).

## ORCID iDs

M.-X. Li  https://orcid.org/0000-0001-5553-2221

W.-H. Deng  https://orcid.org/0009-0001-9669-7222

J. K. H. Tsoi  https://orcid.org/0000-0002-0698-7155

C. Huang  https://orcid.org/0000-0001-9582-7198

## Data Availability

We publicly release the full dataset, benchmark, and the labeling tools at the project website: https://menxli.github.io/metadent/. To protect privacy, all collected images, including both internal clinical images and externally sourced open-access images, have undergone facial anonymization.

# MetaDent: Labeling Clinical Images for
# Vision-Language Models in Dentistry

Meng-Xun Li, Wen-Hui Deng, Zhi-Xing Wu, Chun-Xiao Jin, Jia-Min Wu, Yue Han, James Kit Hon Tsoi, Gui-Song Xia, Cui Huang

## 1. Data Collection and Filtering

### Candidate Mining from COYO-700M

To retrieve intraoral clinical photographs from the large-scale web dataset COYO-700M, we fine-tuned a binary image classifier starting from ViT-L/16 (Dosovitskiy et al. 2020) from pre-trained weights. Positive samples were initially composed of DS1, DS2, MOD (Rashid et al. 2024), and Oral Diseases (Sajid 2023); negatives were drawn from COCO-2017 (Lin et al. 2014) and ImageNet-2012 (Deng et al. 2009). The negatives were down-sampled to match the number of positives to avoid class imbalance, and standard image augmentations (*e.g.,* random crop/resize and horizontal flip) were applied to positives. After the first round, it was observed that the classifier frequently selected face photographs and other non-dental content (*e.g.,* eyes, fingers, ears), and also that many internet images were multi-panel grid collages. We therefore performed an additional multi-class human screening across 19 COYO shards (18,968 images) with five categories: Tooth, Jaw, Face, Oral, and Negative. After de-duplication, 4,487 unique oral images were retained and merged with the positive set; MOD was removed from positives, while Face and Negative were added to the negative pool. We then re-initialized and re-trained the binary classifier from scratch with the refined positive/negative sets, and used it to filter COYO at scale.

Specifically, we fine-tuned a pretrained ViT-L/16 model by replacing the original classification head with a single-output layer for binary classification. All input images were resized to $224 \times 224$ and normalized using the standard ImageNet mean and standard deviation. The entire network (backbone and classification head) was fine-tuned using the Adam optimizer with a learning rate of 1e-6 and the BCEWithLogitsLoss, which integrates a sigmoid activation with binary cross-entropy in a numerically stable manner. The labels were defined as 1 for dental images and 0 for non-dental images, and the classification threshold was set to 0.5. The model was fine-tuned for three epochs and evaluated on an independent test set. And the fine-tuned classifier achieved excellent performance, with a precision of 0.9943, a recall of 0.9961, and an F1-score of 0.9952 for the positive (dental image) class, indicating that it can reliably distinguish dental from non-dental clinical images.

### Handling Collages, Low-Resolution, and Near-Duplicates

To address low-quality web images (grid collages, low resolution), a small detection set was manually annotated using labelImg (exporting in YOLO format), and additional composite images were synthesized together with corresponding masks. Using these data, YOLO11x was fine-tuned to detect collage tiles and automatically crop them into single-panel images. The fine-tuning was performed with an image size of 640, 50 epochs, a batch size of 64, the SGD optimizer, and a close mosaic of 10 to stabilize training during later epochs. This model allowed us to decompose multi-panel collage images into individual intraoral images.

After the collage detection and splitting stage, we further refined the dataset through de-duplication and resolution-based filtering. We then computed average hashes and removed near-duplicates as well as images with short or long edges less than 200 px. The average-hash procedure involved downsampling each image to an 8×8 grayscale representation, computing the mean pixel intensity, and constructing a 64-bit binary hash by thresholding each pixel relative to the mean. This representation is robust to changes in brightness, contrast, and moderate color variation. The Hamming distance between hashes was used to quantify similarity, and images with very small distances were treated as near-duplicates and removed. This hashing-based strategy ensured that the retained images remained visually diverse while redundant or extremely low-quality samples were eliminated.

The resulting collection, combined with DS1 and DS2, constituted the **MetaDent** dataset (total 60,669 images; DS1 after filtering: 4,373; DS2 after filtering: 9,390; DS3 after filtering: 46,906). This multi-source dataset integrates curated clinical images with a large and diverse set of filtered web images, forming the foundation for the downstream benchmarking analyses presented in this study.

## 2. Evaluation Metric

We evaluate VQA performance using the classification accuracy metric, computed as the proportion of questions answered correctly. Because the VQA benchmarks include different question formats, we calculate accuracy separately for multiple-choice questions and for binary true/false questions. In each case, accuracy is defined as:

$$Accuracy = \frac{Number\ of\ correct\ answers}{Total\ number\ of\ questions},$$

applied to the subset of questions of that type. This yields one accuracy score for multiple-choice questions and another for true/false questions, reflecting the model's performance in each format.

For the multi-label classification task, performance is evaluated by comparing the set of predicted labels for each image to the reference label set, deriving standard metrics from the confusion matrix of predictions vs. truth. We count true positives (TP)—the number of labels correctly predicted, false positives (FP)—labels predicted by the model that are not actually present, and false negatives (FN)—labels present in the reference label that the model failed to predict. Using these counts, we compute Precision (P) and Recall (R) as:

$$P = \frac{TP}{TP + FP}, \quad R = \frac{TP}{TP + FN},$$

and the F1-score as the harmonic mean of precision and recall:

$$F_1 = 2 \times \frac{P \times R}{P + R}.$$

In addition to these per-label metrics, we also report a stricter Exact Match criterion for multi-label predictions. Under this metric, a prediction is considered correct only if the entire set of predicted labels exactly matches the set of reference labels for that sample (*i.e.*, no missing or extra labels). The Exact Match score is the percentage of images for which this complete match occurs, and can be expressed as:

$$ExactMatch = \frac{1}{N} \sum_{i=1}^{N} 1\left(Y_{pred}^{(i)} = Y_{true}^{(i)}\right),$$

where $1()$ is an indicator function that equals 1 when the predicted label set $Y_{pred}^{(i)}$ for the image $i$ is identical to the reference set $Y_{true}^{(i)}$, and $N$ is the total number of images. This metric is particularly stringent, as it requires the model to get all labels correct for an image to count as a success.

For the image captioning task, we employ two complementary evaluation methods to assess both the linguistic quality of the captions and their clinical content accuracy:

1. Semantic Similarity (BERTScore): The semantic similarity between each model-generated and reference captions was measured using BERTScore. The reference captions are generated by GPT-OSS to serve as high-quality ground truth descriptions. BERTScore computes token-level semantic similarity using pre-trained contextual embeddings, such as RoBERTa-large (Liu et al., 2019), rather than exact word matches. It outputs precision, recall, and F1 scores based on how well the candidate caption's tokens align with the reference caption's tokens in the embedding space. We report the BERTScore F1, which is the harmonic mean of the precision and recall scores of these semantic matches. This metric reflects how closely the model's captions capture the same semantic meaning and descriptive content as the reference captions.

2. Clinical Content Accuracy (Abnormality Extraction): To evaluate whether the generated captions correctly describe the abnormal findings in the images, GPT-OSS was used to automatically extract the list of abnormalities mentioned in each VLM-generated caption. This yields a set of predicted findings for the image, extracted from the caption text. We then compare this set of extracted abnormalities to the gold-standard set of abnormalities annotated for that image. Treating this as a multi-label comparison, we compute the Precision, Recall, and F1 for the extracted findings by counting true positives (correctly identified abnormalities), false positives (spurious abnormalities mentioned in the caption but not actually present), and false negatives (abnormalities that were missing from the caption). The formulas for P, R, and F1 are analogous to those used in the multi-label classification task above. In this way, the captioning task is evaluated on how accurately it conveys the clinically relevant details, in addition to how fluent or semantically similar it is to a reference description.

All the above metrics are computed on a per data source basis to evaluate performance on each of the three intraoral image sub-datasets individually. We report the metric values for each data source and also provide the overall metrics of the full datasets, giving a measure of the VLM's performance aggregated over the diverse evaluation data.

In addition, for certain VLMs that are unable to provide responses for some of the given intraoral images (*e.g.,* GPT-4o: "I'm sorry, but I can't perform image analysis."; Gemini-2.5-flash: "I'm sorry, but I cannot perform a dental clinical image analysis based on the provided image. The image shows a person's face and neck, but the oral cavity and teeth are not visible. Therefore, I cannot identify any dental conditions or features from the given categories."). These non-responsive cases are excluded from the computation of evaluation metrics (see Appendix Table 13 for details).

## 3. Definitions for Multi-Label Classification and Analysis of Results

The detailed definition for 18-class multi-label classification is presented in Appendix Table 1. Appendix Figures 1–3 present the Precision, Recall, and Macro F1 metrics for each class across datasets and models. Several consistent patterns emerge across all datasets and VLMs:

1. In the F1 heatmaps (Appendix Figure 3), all models exhibit notably lower F1 scores for C2, C3, C13, and C14. Examination of the Precision–Recall plots (Appendix Figures 1–2) reveals a common pattern: precision remains relatively high while recall is low, indicating frequent missed detections.

2. In contrast, C4 shows high recall but low precision in most scenarios, a pattern characteristic of false positives.

These two complementary error modes — under-calling (C2/C3/C13/C14) and over-calling (C4)— are consistently observed across **MetaDent** and all three subsets, reflecting class-specific visual ambiguities that current general-purpose VLM priors fail to model effectively.

To further investigate the domain shift observed in DS2, we compared, for each class, the number of images containing a ground-truth label with the number predicted as positive by each model. Using GPT-4o as a representative example, we identified striking discrepancies in DS2:

1. C4 was predicted in approximately 73.3% of images, whereas only 11% actually contained C4.
2. C5 was predicted in about 37% of images, while only 4% were truly labeled C5.
3. For C14, the most prevalent class in DS2-44.5% of images were truly positive, yet fewer than 9% were predicted positive.

A manual inspection of DS2 revealed two plausible contributing factors:

1. Image tone shift. DS2 images exhibit a noticeable reddish hue, which amplifies soft-tissue redness and mucosal highlights. VLMs may erroneously associate these features with C4/C5—like patterns, leading to false positives.
2. Demographic differences. Variations in patient populations and imaging conditions across datasets may have introduced additional distributional shifts that hinder model generalization. Specifically, DS2 is mostly teenagers with mixed dentition, which differentiates it from the other two datasets.

Finally, the widespread under-detection of C14 across datasets reveals a broader limitation of current VLMs. Malocclusion or dental malalignment demands holistic, structural reasoning across dental arches and occlusal relationships, rather than relying solely on localized texture cues—an ability that current VLMs still lack.

**Appendix Table 1.** Definitions of the 18 categories. Each ID corresponds to two rows: the first row lists the category name, and the second row provides its definition.

| ID | Definition |
|---|---|
| C1 | **Dental caries** |
| | Clearly visible dental caries; early white-spot lesions are excluded. |
| C2 | **Non-carious, unrestored tooth defect** |
| | Refers to tooth fractures or cervical defects not caused by caries and not yet restored (*e.g.,* wedge-shaped defects or notching). Excludes physiological or pathological tooth wear. |
| C3 | **Tooth wear or erosion** |
| | Loss of tooth structure due to physiological or pathological wear, or erosion. Defects caused by caries or minor enamel cracks are excluded. |
| C4 | **Gingival inflammation** |
| | Gingival redness and swelling may present with or without bleeding, and may or may not be accompanied by alveolar bone resorption. |
| C5 | **Gingival recession** |
| | Recession of the gingival margin due to physiological or pathological causes, resulting in root exposure or visible black triangles (interdental gingival recession |
| C6 | **Dental plaque or calculus** |
| | Visible accumulation of plaque or calculus. Excludes occasional food debris. |
| C7 | **Tooth discoloration** |
| | Abnormal tooth color caused by staining, fluorosis, or pulp necrosis, as well as chalky white spots due to enamel demineralization. Excludes dark discoloration due to caries. |
| C8 | **Partial edentulism** |
| | One or more missing teeth with no residual roots present and no prosthetic replacement. |
| C9 | **Residual root** |
| | Complete loss of the clinical crown, with only the root portion remaining. |
| C10 | **Dental filling (direct filling)** |
| | Includes various direct restorative materials on tooth surfaces, such as composite resin, amalgam, temporary fillings, or gutta-percha. |
| C11 | **Fixed prosthesis** |
| | Includes crowns, bridges, veneers, inlays, and other fixed dental prostheses. |
| C12 | **Removable denture** |
| | Includes partial and complete removable dentures. |
| C13 | **Interdental spacing** |
| | Presence of spaces between teeth without missing teeth, possibly due to diastema or physiological spacing. Excludes black triangles caused by gingival recession when adjacent teeth are in contact. |
| C14 | **Malocclusion or dental malalignment** |
| | Includes individual or generalized tooth rotation, crowding, or displacement. The presence of orthodontic appliances does not necessarily indicate malalignment. |
| C15 | **Conventional orthodontic appliance** |
| | Includes brackets, archwires, elastics, and other conventional orthodontic materials. |
| C16 | **Clear aligner orthodontic appliance** |
| | Includes clear aligners, attachments, retainers, and other components of invisible orthodontic systems. |
| C17 | **Oral ulcer** |
| | Includes recurrent aphthous ulcers and traumatic ulcers. Excludes gingival redness or swelling caused by periodontal inflammation. |
| C18 | **Oral wound** |
| | Includes extraction sockets, trauma-related wounds, or surgical wounds of the oral tissues. Excludes gingival redness or bleeding caused by gingivitis |

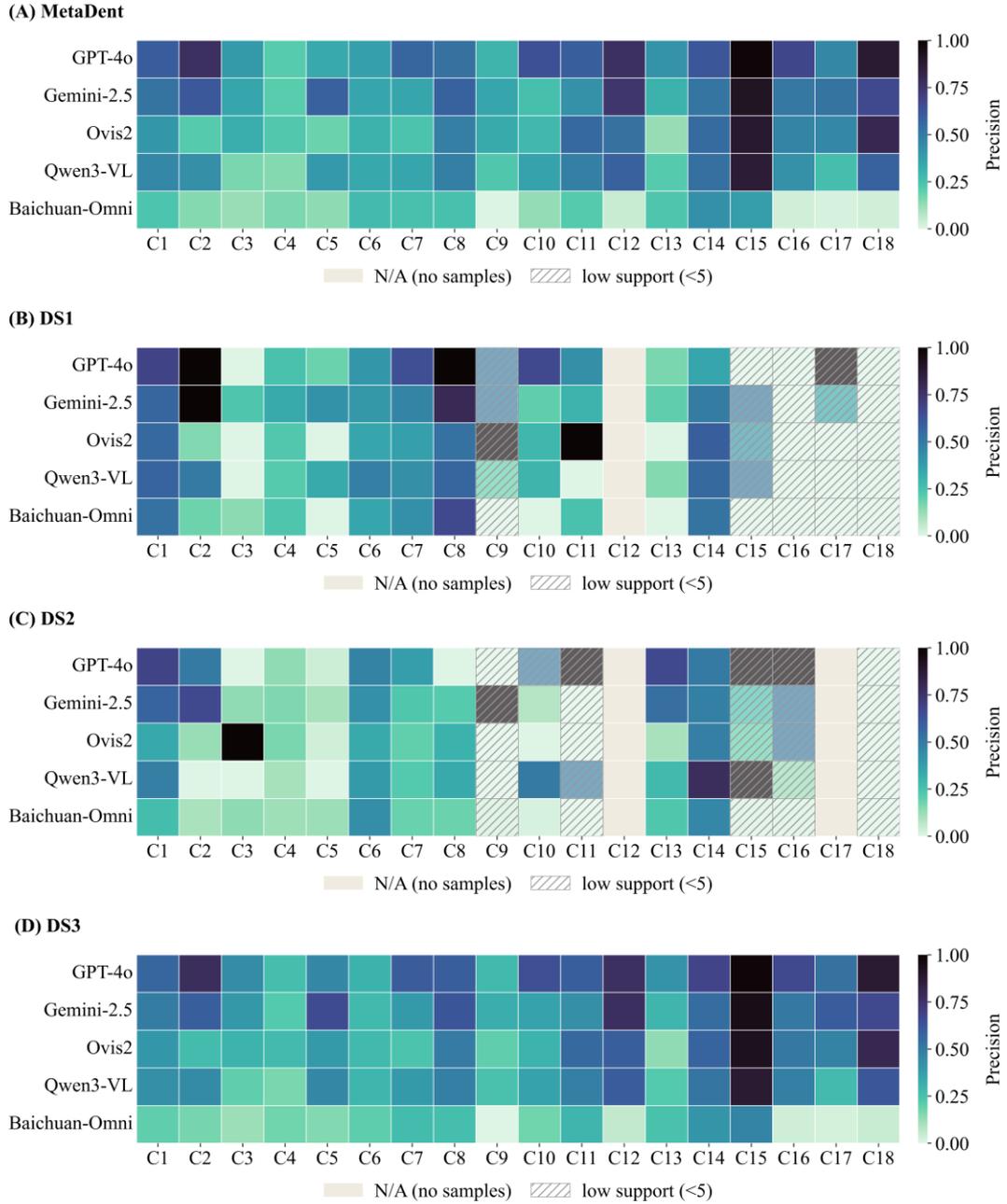

**Appendix Figure 1.** Per-class Precision across datasets. "N/A (no samples)" denotes that the dataset contains no ground-truth instances for that class, and the model did not predict it. "Low support (<5)" flags classes with fewer than five ground-truth samples in the dataset; metrics for these cells should be interpreted with low confidence. (A) MetaDent; (B) DS1; (C) DS2; (D) DS3.

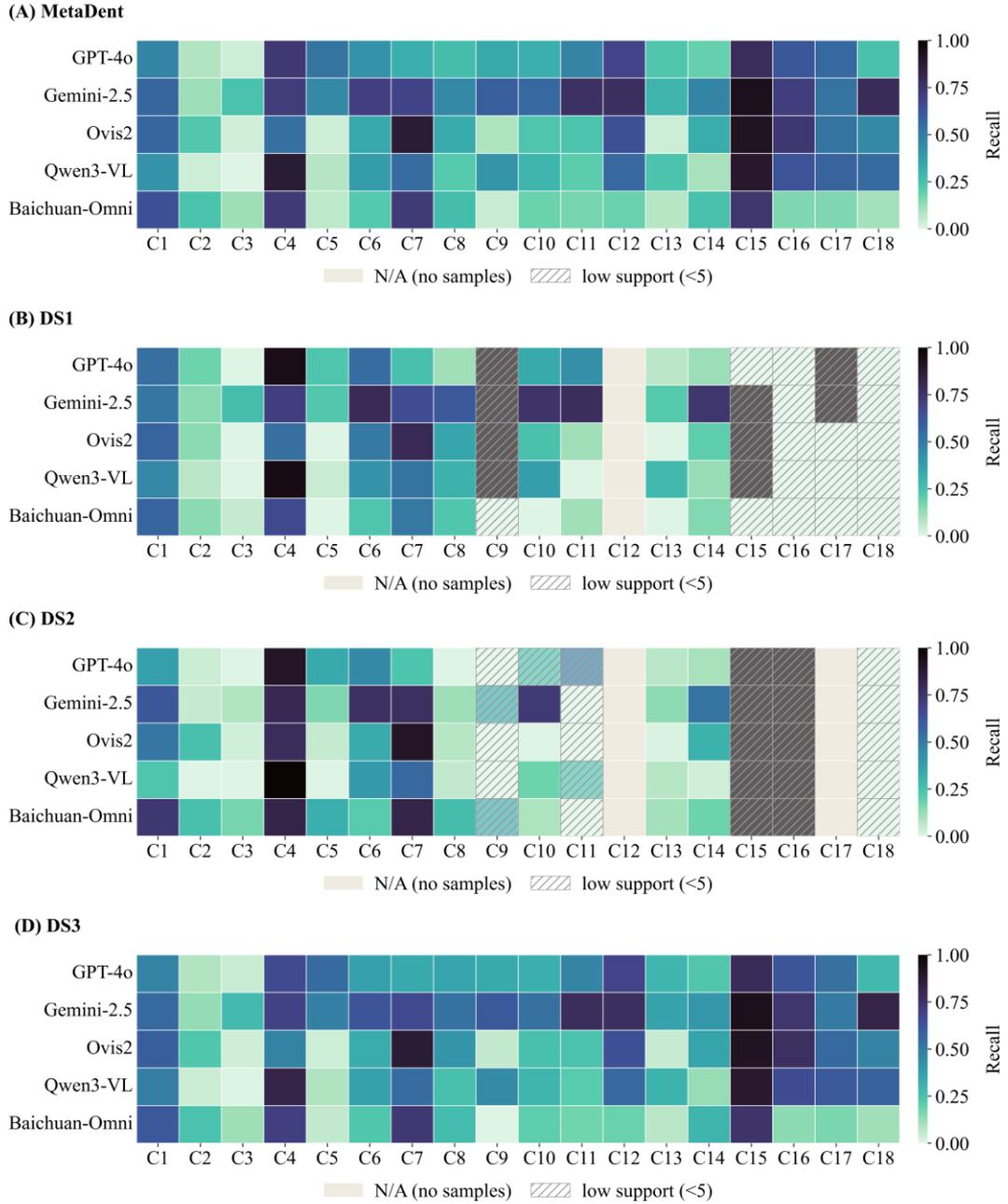

**Appendix Figure 2.** Per-class Recall across datasets. Conventions follow Appendix Figure 1. (A) MetaDent; (B) DS1; (C) DS2; (D) DS3.

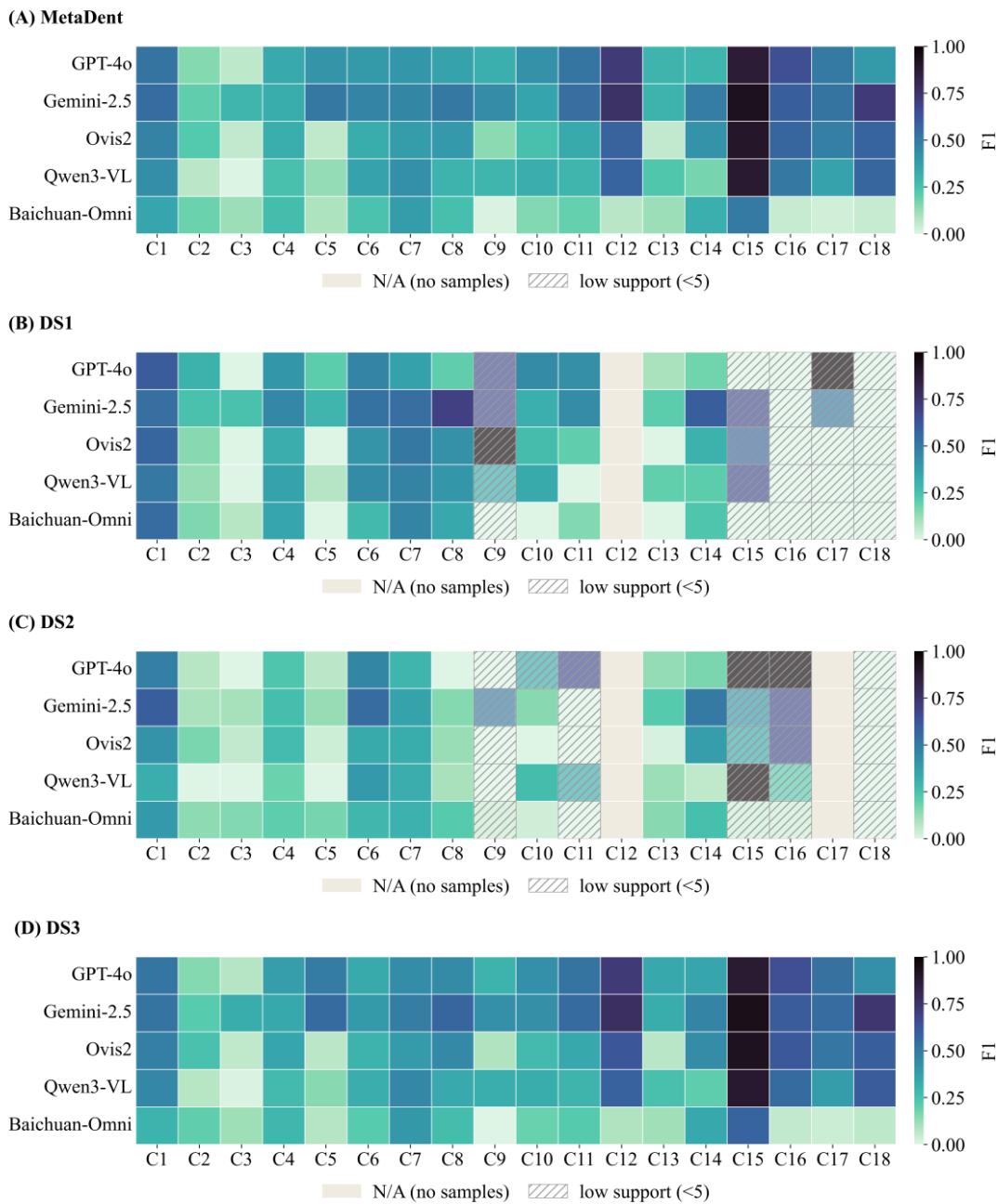

**Appendix Figure 3.** Per-class Macro F1 across datasets. Conventions follow Appendix Figure 1. (A) MetaDent; (B) DS1; (C) DS2; (D) DS3.

# 4.  Errors for Data Conversion

We systematically analyzed the source of errors during secondary dataset generation (see Appendix Figure 4), and detailed definitions for each type of error are presented below.

**Appendix Table 2.** Definition of errors for secondary dataset generation.

| Name | Definition |
|---|---|
| Annotation Error | Errors arising from incorrect or ambiguous labeling in the meta dataset.<br><br>**Example:** *The annotator missed labeling calculus on an image, and the model raised a question, "Is there visible calculus in the image?" with the* |
| | *answer "No".* |
| Tooth Position Error | We found a common pitfall for LLM in reasoning over the meta-label, which is that it cannot accurately capture the spatial relationship between teeth in FDI notation, especially for ranges across the quadrant. This type of error is included in this category.<br><br>**Example:** *The model may fail to recognize #21 is in the range #12-#23. Thus, if the meta label is written in this form, reasoning by LLM will occasionally fail.* |
| Hallucination | The evidence LLM provided supporting its output does not exist on the meta-label, or the evidence exists but is completely irrelevant to its conclusion.<br><br>**Example:** *The model labels an image as "Non-carious unrestored tooth defects" but gives evidence – "The cervical areas of teeth #33, #34 exhibit slight chalky white discoloration", these two are not relevant.* |
| Reasoning Error | The reasoning error can be further categorized into two types:<br>1. Under-reasoning: The LLM fails to draw a valid conclusion that is supported by the evidence in the meta-label.<br>2. Over-reasoning: The LLM generates a proposition that the evidence provided is not sufficient to support, and happens to be wrong for the specific image.<br><br>**Examples:**<br>1. *The meta label clearly mentioned "An image of periodontal surgery" and "bleeding of the gingiva", but the LLM failed to categorize this image into "Oral wound".*<br>2. *On an image with a white spot lesion, the model raises a question: "The small chalky white spots observed on the enamel surfaces are most consistent with which condition?", and gives the answer as "Incipient dental caries (white-spot lesion)". However white spot on the enamel surface alone does not support this proposition.* |
| Translation Error | The original meta-label in this study was annotated in Chinese; in rare circumstances, the translation will contain errors or ambiguities.<br><br>**Example:** *Occasionally, the LLM translates "拉钩" (Hook) into lingual frenum.* |
| Misinterpretation of Definition | This type of error is specific to the task of multi-label classification, where we provide detailed definitions for each category (Refer to Appendix Table 1). If the label provided by the LLM directly violates the definition, it will be classified into this category.<br><br>**Example:** *The model classified an image as "Malocclusion or dental misalignment" given the reason that "Patient wearing a metal orthodontic brace, indicating treatment for malocclusion"; however, the patient in the image has well-aligned teeth, and it is clearly stated in the definition that "The presence of orthodontic appliances does not necessarily indicate malalignment".* |

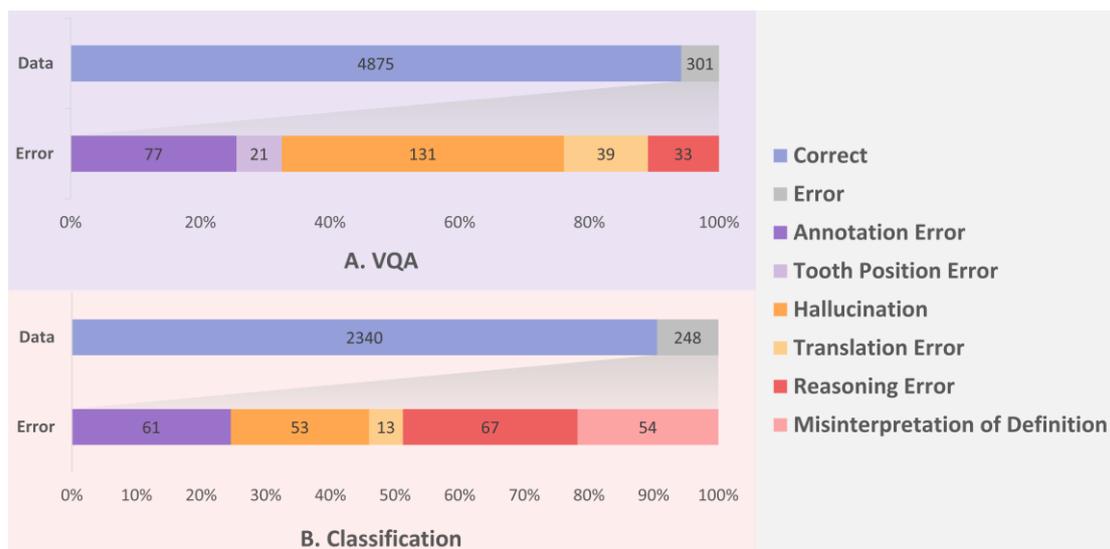

**Appendix Figure 4.** Error analysis of VQA and classification results. (A) Distribution of error types among the 301 incorrect VQA Q-A pairs (out of 5,176 reviewed pairs). (B) Distribution of error types among the 248 classification errors (out of 2,588 reviewed images). Each panel is shown as a horizontal stacked bar chart, with segments colored by error category. Segment labels indicate the raw count of errors of that type and their relative proportion of the total errors in that task.

## 5. Prompt

For each specific task in this study, a fixed prompt template was used consistently across all images to ensure procedural consistency and to minimize prompt-induced variability. For the generation of VQA pairs and multi-label classification labels, the temperature was set to 0 in order to enforce deterministic outputs and maximize reproducibility. For the image captioning task, the temperature was set to 0.6 to allow moderate linguistic diversity while maintaining clinical relevance. In all tasks, each data item was generated only once. However, during VQA generation, if a generated question–answer pair was flagged as inappropriate by the LLM during the built-in self-refinement and self-validation step, the item was automatically regenerated until it satisfied the predefined quality criteria.

During VLM inference, we did not impose an explicit upper bound on the number of generated tokens. All models were allowed to produce outputs until natural completion or until their default system limits were reached. All images were evaluated in their original form without any resizing, normalization, or additional preprocessing. The raw images were directly provided to each VLM according to the default input specifications of the corresponding model API.

For the VQA benchmark, all abnormality entries explicitly flagged as "uncertain" were excluded during question generation by prompt design; that is, the prompts explicitly instructed the LLM not to use uncertain information when constructing questions and answers (as shown in Appendix Table 4). Consequently, no VQA item is based on uncertain annotations. For the multi-label classification task, all final labels were manually reviewed and corrected by dental experts, and no labels marked as uncertain were included in the classification ground truth. In contrast, for the image captioning task, uncertain entries were retained in the reference descriptions. This was done to preserve descriptive richness and to reflect the inherent ambiguity that may exist in real clinical observations, rather than to enforce a strictly deterministic diagnostic target. Notably, the proportion of entries marked as uncertain is small (138/6,452), and therefore, they are unlikely to substantially affect the overall evaluation results.

**Appendix Table 3.** Prompt used for translating meta-labels from Chinese to English.

| Prompt |
| --- |
| You are a professional dentist. Below you will be given a dictionary.<br>Please translate all the values in the dictionary into English, ensuring that the translation is consistent with dental terminology.<br>Finally, keep the output in json format, and do not output any additional content!<br><br>[Input]:<br>\`\`\`json<br>$case<br>\`\`\`<br><br><br>[Output]:<br>Must be output in json format. |

## VQA

**Appendix Table 4.** Prompts to generate VQA from meta-labels.

| Task | Prompt |
| --- | --- |
| Initial VQA generation | You are a professional dentist. Now you are given some descriptive diagnostic texts about a patient's intraoral image JSON format.<br>Please generate visual question answering (VQA) questions based on these texts.<br><br>The question types include:<br>- **$multiple_choice** single-choice questions (each question has four options, A-D, with only one correct answer)<br>- **$true_false** true/false questions (each question has two options, A-B, with only one correct answer)<br><br>[Requirements]:<br>\`\`\`<br>1. All questions must have exactly one correct answer, and the correct answer should be randomly distributed among the options across questions.<br>2. Generate questions only for the content of the image. Do not include any treatment suggestions, or basic dental knowledge.<br>3. Be professional about the wording, carefully review before answering to avoid incorrect dental terminology or descriptions.<br>4. Do not include any question regarding the uncertain content (marked with low_confidence is True in the input text).<br>5. The questions must be concise and precise, avoiding vague or ambiguous wording. No need to mention "based on the description" or "The description mentions"...<br>6. All questions should be generated strictly within the region described in the given text. Any findings not mentioned in the description should be considered normal within that region and may be used to construct questions and answers. For non-existent abnormality questions, you may choose from conditions such as dental caries, non-carious tooth defects, tooth wear or erosion, gingival redness and swelling, gingival recession, dental plaque or calculus, tooth discoloration, dentition defects, residual roots, restorations, fixed prostheses, removable dentures, interdental spacing, dental crowding or malocclusion, traditional orthodontic appliances, clear aligners, or oral ulcers. You may also generate questions based on your other dental knowledge, as long as the content remains consistent with the described region. Any abnormal findings not provided in the |

text should be treated as normal for that region. We require a diverse set of questions. Options may include 'All of the above', 'None of the above' or 'Unknown'.

7. Single-choice questions and true/false questions must test different knowledge points (they cannot be the same or similar). If there is limited abnormal information, you may design questions based on point 6.

8. The output must be in valid JSON array, without any extra content.
```

Note:
When describing tooth positions, the FDI notation is used by default. Sometimes the # symbol is omitted in the description. For example, "11" refers to "Upper Right Central Incisor," and "18" refers to "Upper Right Third Molar.". For ranges, *e.g.,* #12-#23 means "Upper Right Lateral Incisor to Upper Left Canine," which includes #12, #11, #21, #22, and #23.

**FDI Tooth Numbering System**:
| Upper Right | Upper Left |
| Lower Right | Lower Left |

```Permanent tooth
| #18, #17, #16, #15, #14, #13, #12, #11 | #21, #22, #23, #24, #25, #26, #27, #28 |
| #48, #47, #46, #45, #44, #43, #42, #41 | #31, #32, #33, #34, #35, #36, #37, #38 |
```

```Deciduous tooth
| #55, #54, #53, #52, #51 | #61, #62, #63, #64, #65 |
| #85, #84, #83, #82, #81 | #71, #72, #73, #74, #75 |
```

Below is the input JSON:
[Input]:
```json
$case
```

Your output should be a JSON array, where each element is a dictionary containing the following keys:
- "question_type": The question type ID (for single-choice: multiple_choice; for true/false: judge)
- "question": The question text
- "choice": The options, formatted as a dictionary where the key is the option label (A–D for single-choice, A–B for true/false) and the value is the corresponding option text
- "answer": The correct answer (A-D for single-choice; A-B for true/false)
- "reason": The rationale and supporting evidence for why the correct answer is chosen

[Output Template for Each question]:
{
        "question_type": <fill in the question type as specified above>,





<table>
<tr><td></td><td>

"question": <fill in the question text as specified above>,
"choice": <fill in the options as specified above>,
"answer": <fill in the correct answer as specified above>,
"reason": <fill in the rationale and supporting evidence as specified above>
}

[Output Template]:
```json
[
    <fill in the generated questions as specified above>
]
```
</td></tr>
</table>

| Self-verification & regeneration (self-refine) | You are a professional dentist with expertise in oral diagnostics and imaging. You will be provided with two inputs in JSON format: |

You are a professional dentist with expertise in oral diagnostics and imaging. You will be provided with two inputs in JSON format:
- A descriptive diagnostic text detailing clinical findings from a patient's intraoral images.
- An AI-generated Visual Question Answering (VQA) set, consisting of questions and answers about the same images.
Your task is to evaluate the consistency of the AI-generated VQA content against the diagnostic text and revise it as necessary to ensure full alignment.

If you determine that the current VQA does not match the texts, you must revise the VQA so that it satisfies the requirements.
Note that:
1. The question must focus on the region described in the text or using the region inferred from the description. For example, an image showing only the upper jaw should not have questions about the lower jaw (or the answer should be "Unknown" if the question is about the lower jaw, or the question should be about the visibility of the lower jaw).
2. Any abnormalities/disease not mentioned in the text (but within the described region) are assumed absent, constructing questions based on such deduced non-existent abnormalities is acceptable.
3. If any item in the diagnostic text is marked with low_confidence: true, it must not be used to generate or support any question or answer. Any VQA content relying on such uncertain observations must be revised or removed.
4. Be conservative in your judgement and revisions. If you are unsure, it is better to mark the VQA as valid. If the VQA content is largely correct with only minor issues, make minimal necessary changes to ensure accuracy and alignment with the text.
5. When describing tooth positions, the FDI notation is used by default. Sometimes the # symbol is omitted in the description. For example, "11" refers to "Upper Right Central Incisor," and "18" refers to "Upper Right Third Molar.". Interpret tooth ranges as inclusive sequences: ranges spanning opposite quadrants (*e.g.,* 12–22 or 22–12) include all teeth crossing the midline (12,11,21,22); ranges within the same quadrant (*e.g.,* 21–25) include sequential teeth in that quadrant (21,22,23,24,25).

**FDI Tooth Numbering System**:
| Upper Right | Upper Left |
| Lower Right | Lower Left |

```Permanent tooth

| #18, #17, #16, #15, #14, #13, #12, #11 | #21, #22, #23, #24, #25, #26, #27, #28 |
| #48, #47, #46, #45, #44, #43, #42, #41 | #31, #32, #33, #34, #35, #36, #37, #38 |
```

```Deciduous tooth
| #55, #54, #53, #52, #51 | #61, #62, #63, #64, #65 |
| #85, #84, #83, #82, #81 | #71, #72, #73, #74, #75 |
```

Below is the patient's descriptive diagnostic text:
[Diagnostic Text Input]:
```json
$case
```

Below is the AI-generated VQA content:
[AI Input]:
```json
$ai_input
```

Your output should be a JSON object with the following keys:
- "invalid": Whether there is an error with the current VQA content. (true/false)
- "error_type": One of the following error categories:
    * multiple answers present
    * using uncertain information
    * incorrect answer
    * common knowledge error
    * original annotation insufficient detail
    * out-of-region question
    * incorrect tooth position
    * hallucination (error within the described region)
    * other
    * null (if "invalid" is false)
- "evidence": The evidence text from the input that supports your judgment.
- "new_question": A dictionary containing the corrected VQA data. This field is required only when "invalid" is true.

Note on construction of the revised question if new_question is needed (besides the above criteria):
The question should be a JSON array, where each element is a dictionary containing the following keys:
- "question_type": The question type ID (for single-choice: multiple_choice; for true/false: judge)
- "question": The question text
- "choice": The options, formatted as a dictionary where the key is the option label (A–D for single-choice, A–B for true/false) and the value is the corresponding option text
- "answer": The correct answer (A-D for single-choice; A-B for true/false)
- "reason": The rationale and supporting evidence for why the correct answer is





chosen

1. Generate questions only for the content of the image. Do not include any treatment suggestions, or basic dental knowledge.

2. Be professional about the wording, carefully review before answering to avoid incorrect dental terminology or descriptions.

3. The questions must be concise and precise, avoiding vague or ambiguous wording. No need to mention "based on the description" or "The description mentions"...

4. All questions should be generated strictly within the region described in the given text. Any findings not mentioned in the description should be considered normal within that region and may be used to construct questions and answers. For non-existent abnormality questions, you may choose from conditions such as dental caries, non-carious tooth defects, tooth wear or erosion, gingival redness and swelling, gingival recession, dental plaque or calculus, tooth discoloration, dentition defects, residual roots, restorations, fixed prostheses, removable dentures, interdental spacing, dental crowding or malocclusion, traditional orthodontic appliances, clear aligners, or oral ulcers. You may also generate questions based on your other dental knowledge, as long as the content remains consistent with the described region. Any abnormal findings not provided in the text should be treated as normal for that region. We require a diverse set of questions. Options may include 'All of the above', 'None of the above' or 'Unknown'.

The output must be in JSON format.
[Output Template]:
```json
{
    "invalid": true/false,
    "error_type": ...,
    "evidence": ...,
    "new_question": ...
}
```

**Appendix Table 5.** Prompt for VLMs to answer VQA.

| Prompt |
| --- |
| You are a professional dentist. You are now presented with a clinical image of a patient and a multiple-choice question.<br>Please select only one correct answer based on the visual evidence from the image.<br><br>Below is the question:<br>[Question]:<br>```json<br>{<br>    "question": $question,<br>    "choice": $choice<br>}<br>```<br><br>Your output should be a JSON object with the following keys:<br>- "answer": Your selected option, represented as one of $answer_options.<br>- "reason": The reasoning and supporting visual evidence for your chosen answer. |





Do not include any additional explanations, text, or formatting outside the JSON.

[Output Template]:
```json
{
    "answer": <fill in the selected answer as specified above>,
    "reason": <fill in the reasoning and supporting evidence as specified above>
}
```

## Classification

**Appendix Table 6.** Prompts for extracting multi-class categories.

| Task | Prompt |
|---|---|
| Extraction based on abnormal text annotations | You are a professional dentist. Now you have some descriptive diagnostic texts about patients in JSON format.<br>Please perform multi-class category extraction based on these texts.<br>The categories and some extra information are as follows:<br><br>[Categories]:<br>$label_desc<br><br>Below is the input JSON:<br>[Input]:<br>```json<br>$case<br>```<br><br><br>Your output should be a JSON array, where each element is a dictionary containing the following keys:<br>- "id": The category ID (*e.g.,* "C1", "C2", etc.)<br>- "name": The category name<br>- "evidence": The evidence text from the input that supports the classification into this category.<br><br>You can do some basic inference based on the input text to provide the evidence.<br>But if the input does not contain information related to a specific category, do not include that category in the output.<br>Additionally, if low_confidence is True, do not include this annotation in the multi-class extraction.<br><br>[Output Template]:<br>```json<br>[<br>    <fill in the extracted categories as specified above><br>]<br>``` |
| Extraction based on images | You are a professional dentist. You are now given a clinical image of a patient.<br>Please perform multi-class category extraction based on this dental clinical image. |





The categories and additional information are as follows:
[Categories]:
$label_desc

Your output should be a JSON array, where each element is a dictionary containing the following keys:
- "id": The category ID (*e.g.,* "C1", "C2", etc.)
- "name": The category name
- "evidence": The evidence or visual cues observed in the image that support the classification into this category.

Important requirements:
- Only select categories that are visibly present in the image. Do not select or provide explanations for categories that cannot be seen.
- The "id" and "name" must strictly match and correspond to the given categories.
- You must only choose from the listed categories.
- It is acceptable to output an empty array if no categories apply.

[Output Template]:
```json
[
    <fill in the extracted categories as specified above>
]
```

## Image Captioning

**Appendix Table 7.** Prompts for generating image captions.

| Task | Prompt |
|------|--------|
| Generated from abnormal text annotations | You are a professional dentist. You are now given descriptive diagnostic texts in JSON format about a patient's intraoral image.<br><br>Based on these texts, please generate a detailed description of the image content. Please note:<br>- You are can be imaginative, assuming non-mentioned common abnormalities are normal for that region, you are encouraged to describe about these deduced normal content as well.<br>- While you are encouraged to be imaginative, use your imagination with reasoning based on provided content, ensuring that all your output is not against the input.<br>- You must respond in (un-structured) natural language, instead of structured formats like bullet points or numbered lists.<br>- Adjust your tongue as if you are seeing the image instead of respond based on the text input.<br><br>Below is the input JSON:<br>[Input]:<br>```json<br>$case<br>``` |





| | |
|---|---|
| | Your output should be a JSON object with the following keys:<br>- "description": A comprehensive description generated from the given texts.<br><br>[Output Template]:<br>```json<br>{<br>    "description": "<fill in the detailed description including observed content, dental instruments, and abnormal findings>"<br>}<br>```<br>Please output json directly without extra output. |
| Generated from images | You are a professional dentist. You are now given a clinical image of a patient. Please generate a detailed and vivid natural language description based on this image.<br><br>Your output should be a JSON object with one key, "description". The description must be written as a coherent paragraph, not as a list or dictionary. It should clearly and naturally describe the imaging direction (the angle or orientation from which the image was captured), the main subject of the image (the primary anatomical focus or structure), and all observed abnormalities (including pathological findings, dental defects, or visible dental instruments related to these abnormalities). You may also describe regions that appear normal if you are confident in their correctness, but your descriptions must remain accurate and factual, without any fabricated or speculative details. Use multiple sentences as needed to make the description fluent, expressive, and clinically meaningful.<br><br>[Output Template]:<br>```json<br>{<br>    "description": "<fill in the detailed description including observed content, dental instruments, and abnormal findings>"<br>}<br>```<br>Please output json directly without extra output. |

**Appendix Table 8.** Prompts for extracting abnormal findings from the generated image captions.

| Prompt |
|---|
| You are a professional dentist. You are now given descriptive diagnostic texts in JSON format about a patient's intraoral image.<br><br>Based on the given texts, please list all observed abnormalities and present them in a list format.<br><br>Below is the input JSON:<br>[Input]:<br>```json<br>$case<br>```<br><br>Definition of abnormality: Any pathological findings, dental defects, or visible dental instruments associated with abnormalities. Do not describe normal findings.<br><br>Your output should be a JSON array, where each element is a dictionary containing the following |





keys:
- "abnormality": A summarized description of the abnormality based on the given texts
- "reason": The reasoning and supporting evidence for this abnormality description

[Output Template]:
```json
[
    <fill in the extracted abnormalities as specified above>
]
```

**Appendix Table 9.** Prompts for computing the image captioning confusion matrix.

| Prompt |
| --- |
| You are a professional dentist. You are now given two sets of information in JSON format about a patient's intraoral image: Reference – the ground truth abnormality descriptions. Prediction – the abnormality descriptions generated by an LLM.<br><br>Below is the Reference JSON input:<br>[Reference]:<br>```json<br>$reference<br>```<br><br>Below is the Prediction JSON input:<br>[Prediction]:<br>```json<br>$prediction<br>```<br><br>Definition of abnormality: Any pathological findings, dental defects, or visible dental instruments associated with abnormalities.<br><br>Based on these two inputs, please calculate the values for the confusion matrix: True Positive (TP), False Negative (FN), False Positive (FP), and True Negative (TN).<br><br>Your output should be a JSON object with the following keys:<br>- "TP": integer value<br>- "FN": integer value<br>- "FP": integer value<br>- "TN": integer value<br>- "reason": the reasoning and supporting evidence for these values<br><br>[Output Template]:<br>```json<br>{<br>    <fill in the confusion matrix values as specified above><br>}<br>``` |

# 6. Examples of common failure modes

To complement the quantitative evaluation, we conducted a qualitative analysis to illustrate





common failure modes of current VLMs in intraoral image understanding. While some AI-generated descriptions demonstrate correct recognition of prominent dental structures and overall oral health status, they frequently omit or misinterpret clinically relevant abnormalities when multiple findings coexist within the same image.

As shown in Appendix Figure 5, the VLM-generated description correctly identifies some salient features; however, several interpretations are incorrect. For example, the statement that the "enamel looks intact" is inaccurate. In addition, the VLM does not capture some of the other abnormalities present in the image.

These examples highlight a key limitation of current VLMs: although they can generate fluent and seemingly plausible descriptions, their understanding often remains incomplete at the level required for reliable clinical interpretation. Such failure modes help explain why relatively low quantitative scores (e.g., F1 or abnormality-level recall) correspond to meaningful gaps in clinical adequacy.

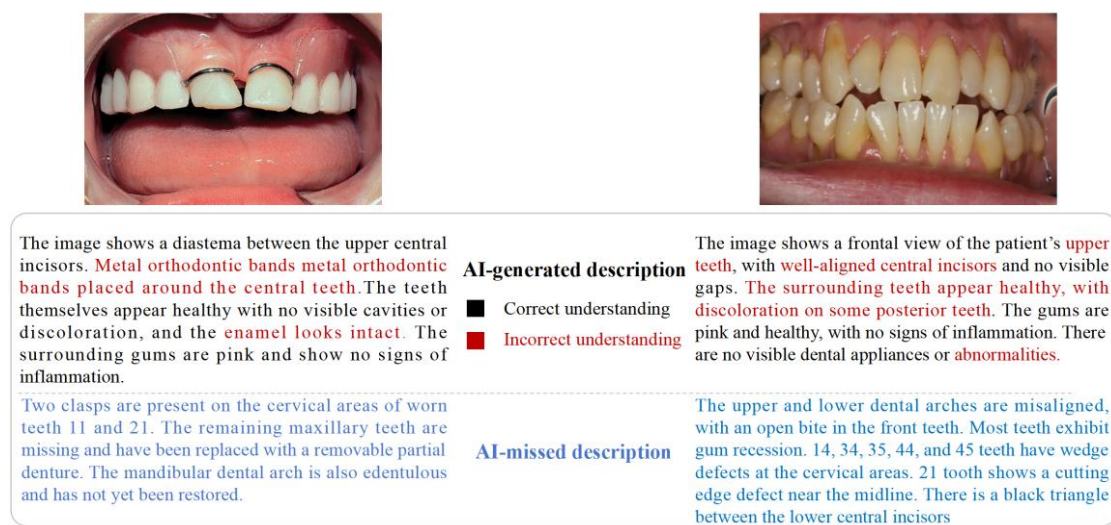

**Appendix Figure 5.** Representative examples of AI-generated image descriptions illustrating common failure modes in intraoral image understanding. Correctly recognized findings are highlighted in black, while incorrect interpretations are shown in red. Blue text indicates clinically relevant findings missed by the AI model according to expert annotations. These examples demonstrate that, despite fluent language generation, VLMs may overlook or misinterpret subtle but important dental abnormalities.

# 7. Advantages of the Labeling Scheme

Our annotation protocol was designed to reflect the nuanced reasoning and descriptive richness inherent in clinical dental practice. Rather than constraining annotators to a fixed set of categorical labels, we instructed annotators to describe intraoral findings using natural, unstructured language as they would in a clinical note. This approach enables the capture of hierarchical, multi-scale, and often interdependent conditions that cannot be adequately represented by flat or discrete classification schemes. Below are two concrete cases to help readers establish a conceptual foundation of our annotation methodology:

## Diagnostic Inference in Annotations

One desirable property of labeling with free text is that we can label beyond what we see. Specifically, when the visual findings in the image clearly (or with high probability) supported a specific clinical diagnosis or indicated a prior or planned clinical intervention, the annotators (clinicians) are allowed to label beyond pure descriptions and supplement their descriptive observations with extrapolated diagnostic inferences.





Given the following image:

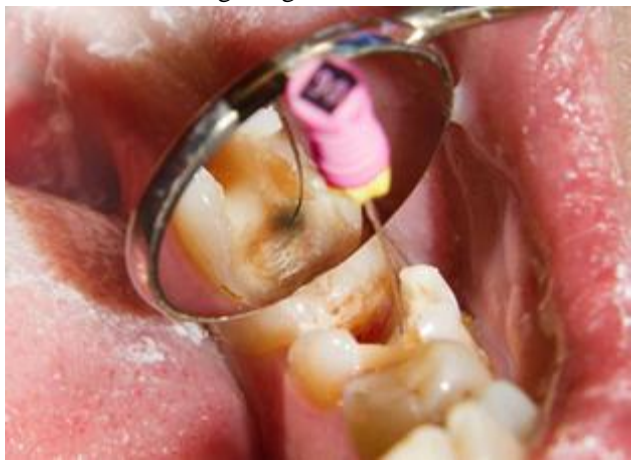

What can be observed is that the mandibular left first molar (tooth #36) exhibits extensive crown destruction. A root canal file is inserted into the remaining tooth structure, and the surrounding dentin is darkened (though it still appears hard).

Although the image does not show extensive carious lesions, it is reasonable to infer that the tooth previously had significant caries, which has since been excavated. In this case, the annotators may label this tooth as: "Tooth #36 has extensive post-carious crown destruction; root canal file in situ; surrounding dentin appears dark but sound." Of which the "extensive post-carious crown destruction" is extrapolation by clinical experience, since no significant caries are observed on the tooth (excavated).

## Capturing the Hierarchical Complexity

Another desirable property of labeling with free text is that we can label beyond predefined categories. Traditional approaches typically assign each tooth or lesion a single, discrete label from a predefined set. While useful for basic classification tasks, such schemes are inherently limited by disregarding clinically relevant contextual details, e.g., it is hard to convey spatial distributions, severity gradients, or rare cases. Our narrative approach preserves this complexity.

For example, consider this intraoral image showing teeth #15–17:

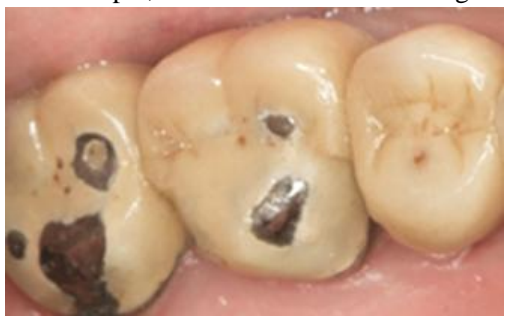

With conventional categorical labels, one may be forced to label the teeth as "PFM crown" or perhaps with a "Defective restoration" or "Ceramic chipping" to #16 and #17, if a hierarchical label scheme is adopted.

In contrast, the same region can be annotated with rich, clinically grounded descriptions. Perhaps as a whole:





➢ "Tooth #15 is a porcelain crown. Teeth #16 and #17 are porcelain-fused-to-metal (PFM) crowns, with metal exposure on the occlusal surfaces and cervical margins; tooth #17 also has ceramic chipping on the palatal side."

Or maybe separately as:
➢ "Porcelain full crown" for #15
➢ "PFM crown with metal exposure on occlusal and cervical regions" for #16
➢ "PFM crown with ceramic chipping on occlusal and palatal surfaces, and metal exposure at cervical margins. A white spot can be observed on the mesial occlusal metal surface, presumably a perforation" for #17

Notably, this flexible framework allows us to use concise labels for common, straightforward cases while reserving detailed narrative descriptions for complex, rare, or ambiguous findings. This design aligns closely with clinical reasoning patterns.

Meanwhile, this method is flexible and interoperable with existing AI workflows. Whichever labeling methods (holistically or separately) we use, they retain sufficient semantic richness to support diverse downstream applications. If the target application requires a restricted categorial label set, we can use an LLM to convert these meta-labels to the same as conventional categorical labels for subsequent use (as we did for the classification task in the manuscript). Moreover, the annotation is inherently caption-friendly: because the labels are rich in descriptive semantics and structured in natural language, they can be readily adapted—without loss of information—into fluent, image-aligned captions for vision–language tasks.

## 8. Application of Dental-specific VLMs

The clinical image vision-language data developed in this project can serve as a foundational infrastructure for advancing AI models in dentistry. The potential impacts of the VLMs can be outlined across three key dimensions—clinical care, public health, and the evolution of AI technologies, as detailed below:

1. **Enhance clinical workflow:** VLMs can be integrated into digital clinical platforms in dental clinics or hospitals. Such integration may enable automatic semantic interpretation and structured documentation of intraoral images, reducing clinicians' manual burden in composing electronic health records (EHRs). This improves both the efficiency and standardization of clinical documentation.

2. **Promote public health:** In primary care settings or resource-constrained regions, the dental-specific VLMs can function as an intelligent preliminary screening tool. It can assist non-dental healthcare providers—such as general practitioners, community health workers, or public health personnel—in the initial identification and risk assessment of common oral diseases, thereby helping bridge gaps caused by shortages of dental specialists. Moreover, the high-performing and semantically precise VLM can provide quantifiable and traceable data to support regional or nationwide epidemiological studies of oral diseases, assessments of disease burden, and monitoring of public health intervention outcomes.

3. **Cross-modal representation learning**: The fine-grained vision–language data pairs will significantly accelerate research in cross-modal representation learning within dentistry. On one hand, they enable semantic-level intelligent retrieval from dental image databases, thereby enhancing research and educational efficiency. On the other hand, these structured textual descriptions can be leveraged to train generative models for high-quality synthetic data generation—either visual or textual—thus alleviating challenges posed by medical data scarcity and privacy constraints. Such resources can serve as critical training assets for next-generation dental foundation models and multimodal diagnostic agents.





## 9. Cross-LLM Validation for Potential Benchmark Bias

The detailed Precision and Recall results for the original multi-label Classification (CLS) and Image Captioning (CAP) tasks are summarized in Appendix Table 10.

For the CLS task, all models exhibit moderate precision and recall, with clear performance variation across data sources. Overall, GPT-4o achieves the highest average precision (0.475) and recall (0.438), indicating a relatively balanced ability to identify abnormalities while controlling false positives. Gemini-2.5 shows a recall-oriented behavior, achieving the highest average recall (0.444) but with slightly lower precision (0.398). In contrast, the open-source models (Qwen3-VL, Ovis2, and Baichuan-Omni) demonstrate lower and less balanced precision–recall profiles, reflecting either missed findings or increased false-positive predictions. Across models, performance on DS2 is consistently lower than on DS1 and DS3, further highlighting the domain shift present in this subset.

For the CAP task, both precision and recall remain substantially lower than those observed in classification. Across all models and datasets, average precision ranges from 0.093 to 0.148 and recall from 0.072 to 0.140, indicating that most clinically relevant abnormalities are either incompletely described or omitted in free-form captions. Gemini-2.5 attains the highest average precision (0.148) and recall (0.140), while GPT-4o and Ovis2 tend to produce more conservative captions with relatively lower recall. These results suggest that, although current VLMs can generate fluent textual descriptions, their ability to consistently capture clinically salient findings in captioning remains limited.

**Appendix Table 10.** Precision (P) and recall (R) results for the multi-label classification (CLS) and image captioning (CAP) tasks across different data sources (DS1, DS2, DS3) and the overall MetaDent average (Avg). The results show consistent performance differences across models and data sources, with notably lower scores on DS2, reflecting the presence of domain shift.

| VLMs | Score | DS1 | DS2 | DS3 | Avg |
|---|---|---|---|---|---|
| GPT-4o | CLS-P | 0.493 | 0.243 | 0.519 | 0.475 |
| | CLS-R | 0.453 | 0.238 | 0.475 | 0.438 |
| | CAP-P | 0.091 | 0.152 | 0.103 | 0.112 |
| | CAP-R | 0.053 | 0.067 | 0.076 | 0.072 |
| Gemini-2.5 | CLS-P | 0.449 | 0.295 | 0.418 | 0.398 |
| | CLS-R | 0.493 | 0.435 | 0.441 | 0.444 |
| | CAP-P | 0.122 | 0.130 | 0.156 | 0.148 |
| | CAP-R | 0.101 | 0.106 | 0.153 | 0.140 |
| Qwen3-VL | CLS-P | 0.415 | 0.197 | 0.384 | 0.351 |
| | CLS-R | 0.397 | 0.197 | 0.349 | 0.325 |
| | CAP-P | 0.083 | 0.072 | 0.099 | 0.093 |
| | CAP-R | 0.081 | 0.077 | 0.100 | 0.094 |
| Ovis2 | CLS-P | 0.468 | 0.237 | 0.359 | 0.345 |
| | CLS-R | 0.469 | 0.284 | 0.336 | 0.337 |
| | CAP-P | 0.113 | 0.107 | 0.107 | 0.108 |
| | CAP-R | 0.100 | 0.080 | 0.083 | 0.084 |
| Baichuan-Omni | CLS-P | 0.436 | 0.177 | 0.348 | 0.323 |
| | CLS-R | 0.461 | 0.264 | 0.373 | 0.360 |
| | CAP-P | 0.082 | 0.087 | 0.100 | 0.096 |
| | CAP-R | 0.080 | 0.081 | 0.089 | 0.086 |

Since large language models (LLMs) are involved in both benchmark construction and evaluation





in this study, a potential concern is the risk of circularity and linguistic prior bias, particularly for the image captioning task in which an LLM is used to extract abnormal findings for clinical consistency assessment. To examine whether such bias could materially affect the benchmark outcomes, we conducted an additional cross-LLM validation experiment.

Specifically, in addition to the original captioning benchmark generated using GPT-OSS, we re-generated the entire reference caption dataset using an independent LLM, Qwen3-Next-80B-A3B-Instruct, while keeping all other components of the evaluation pipeline unchanged. All five VLMs were then re-evaluated on both captioning benchmarks. The comparative results are reported in Appendix Tables 11 and 12.

Across all three data subsets (DS1, DS2, and DS3) and for all evaluated VLMs, the relative performance rankings and overall difficulty levels remained stable between the GPT-OSS–generated and Qwen-generated benchmarks. No systematic advantage was observed for VLMs that share a similar linguistic family with the reference caption generator. In fact, a modest overall performance increase was observed when using the Qwen-generated references, suggesting that the benchmark is not biased in favor of any specific LLM family.

**Appendix Table 11.** Semantic consistency of VLM-generated captions evaluated using BERTScore-F1 on the captioning benchmarks generated by GPT-OSS and Qwen, respectively. The results indicate that no evident benchmark bias is observed for either GPT-based or Qwen-based VLMs.

| VLMs | Score | DS1 | | DS2 | | DS3 | | Avg | |
|---|---|---|---|---|---|---|---|---|---|
| | | GPT | Qwen | GPT | Qwen | GPT | Qwen | GPT | Qwen |
| GPT-4o | F1 | 0.174 | 0.195 | 0.210 | 0.231 | 0.204 | 0.221 | 0.203 | 0.221 |
| Gemini-2.5 | F1 | 0.184 | 0.198 | 0.193 | 0.213 | 0.216 | 0.222 | 0.209 | 0.218 |
| Qwen3-VL | F1 | 0.177 | 0.197 | 0.180 | 0.201 | 0.216 | 0.223 | 0.206 | 0.217 |
| Ovis2 | F1 | 0.115 | 0.145 | 0.138 | 0.162 | 0.164 | 0.188 | 0.155 | 0.180 |
| Baichuan-Omni | F1 | 0.099 | 0.123 | 0.114 | 0.132 | 0.137 | 0.152 | 0.129 | 0.146 |

**Appendix Table 12.** Clinical consistency of VLM-generated captions was evaluated by extracting abnormal findings from the captions using an LLM and comparing them with the ground-truth abnormalities, based on captioning benchmarks generated by GPT-OSS and Qwen. The results show no evident benchmark bias for either GPT-based or Qwen-based VLMs.

| VLMs | Score | DS1 | | DS2 | | DS3 | | Avg | |
|---|---|---|---|---|---|---|---|---|---|
| | | GPT | Qwen | GPT | Qwen | GPT | Qwen | GPT | Qwen |
| GPT-4o | P | 0.091 | 0.104 | 0.152 | 0.157 | 0.103 | 0.131 | 0.112 | 0.134 |
| | R | 0.053 | 0.056 | 0.067 | 0.079 | 0.076 | 0.094 | 0.072 | 0.088 |
| Gemini-2.5 | P | 0.122 | 0.122 | 0.130 | 0.146 | 0.156 | 0.185 | 0.148 | 0.173 |
| | R | 0.101 | 0.112 | 0.106 | 0.127 | 0.153 | 0.182 | 0.140 | 0.166 |
| Qwen3-VL | P | 0.083 | 0.104 | 0.072 | 0.091 | 0.099 | 0.111 | 0.093 | 0.107 |
| | R | 0.081 | 0.091 | 0.077 | 0.097 | 0.100 | 0.120 | 0.094 | 0.113 |
| Ovis2 | P | 0.113 | 0.091 | 0.107 | 0.097 | 0.107 | 0.113 | 0.108 | 0.109 |
| | R | 0.100 | 0.075 | 0.080 | 0.080 | 0.083 | 0.091 | 0.084 | 0.088 |
| Baichuan-Omni | P | 0.082 | 0.084 | 0.087 | 0.086 | 0.100 | 0.123 | 0.096 | 0.113 |
| | R | 0.080 | 0.076 | 0.081 | 0.084 | 0.089 | 0.110 | 0.086 | 0.103 |

## 10.Statistical Analysis

### Annotation Consistency Assessment

To assess the consistency between the two annotators, we evaluated inter-rater reliability using Cohen's Kappa. Each image–label combination was treated as an independent sample. We first





randomly selected 100 images from the dataset and had them independently annotated by both raters. The annotated abnormality descriptions were then converted into multi-label classification vectors using an LLM, with each image represented as an 18-dimensional binary array (1 indicating the presence of a category and 0 indicating its absence), and agreement was evaluated across all label assignments. The resulting Kappa coefficient was 0.83, indicating a high level of agreement between annotators.

## Statistical Significance Analysis of VLM Performance

To assess whether the five evaluated VLMs differed significantly in their performance, we conducted statistical significance analyses on the two tasks where binary success–failure outcomes can be clearly defined: VQA and multi-label classification. For both tasks, the analyses were restricted to the subset of items for which all five VLMs produced valid responses, thereby avoiding bias introduced by non-responses or output-format errors.

The hypotheses for the Cochran's Q test were defined as follows:

1) Null hypothesis ($H_0$): All models have equal probabilities of correct prediction.
2) Alternative hypothesis ($H_1$): At least one model has a different probability of correct prediction.

For the VQA task, Cochran's Q test revealed highly significant differences among the five models ($Q = 225.99$, p = 9.66e-48), indicating that the VLMs do not share the same probability of generating correct answers. To further investigate the pairwise differences, McNemar tests with Bonferroni correction were applied. The results (see Appendix Table 14 and Appendix Figure 6A) demonstrate that numerous model pairs exhibit statistically significant differences, with GPT-4o, Gemini-2.5, and Qwen3-VL showing several strong contrasts. These findings confirm that VQA performance varies substantially across current VLMs.

For the multi-label classification task, Cochran's Q test again demonstrated significant model-level differences ($Q = 181.78$, p = 3.09e-38). Pairwise McNemar tests with Bonferroni correction (see Appendix Table 15 and Appendix Figure 6B) show that GPT-4o consistently outperforms most other models, while the differences among Gemini-2.5, Qwen3-VL, and Baichuan-Omni are less pronounced.

No significance testing was conducted for the image captioning task because caption correctness does not map cleanly to a binary outcome and lacks a well-defined physical or clinical interpretation for hypothesis testing. Instead, captioning quality is evaluated through semantic similarity (BERTScore) and abnormality-level precision/recall metrics, which are presented descriptively in the manuscript.

**Appendix Table 13.** Percentage of non-responses for each VLM across the VQA, multi-label classification (CLS), and image captioning (CAP) tasks.

| VLMs | VQA | CLS | CAP |
|---|---|---|---|
| GPT-4o | 8.58% | 25.46% | 2.32% |
| Gemini-2.5 | 0.18% | 0.04% | 0.00% |
| Qwen3-VL | 1.90% | 0.00% | 0.00% |
| Ovis2 | 0.02% | 0.00% | 0.00% |
| Baichuan-Omni | 0.00% | 0.12% | 0.00% |

**Appendix Table 14.** Pairwise McNemar test results with Bonferroni correction for the VQA task. Variable b denotes the number of samples correctly predicted by Model 1 but incorrectly predicted by Model 2, and c denotes the number of samples incorrectly predicted by Model 1 but correctly predicted





by Model 2. An asterisk (*) indicates a statistically significant difference. A Bonferroni-corrected p-value less than 0.005 was considered statistically significant.

| Model 1 | Model 2 | b | c | p-value |
|---|---|---|---|---|
| GPT-4o | Gemini-2.5 | 1,626 | 2,081 | 8.87e-14* |
| GPT-4o | Qwen3-VL | 2,348 | 2,072 | 3.53e-05* |
| GPT-4o | Ovis2 | 2,284 | 2,422 | 0.0458 |
| GPT-4o | Baichuan-Omni | 3,003 | 2,511 | 3.79e-11* |
| Gemini-2.5 | Qwen3-VL | 2,407 | 1,676 | 3.16e-30* |
| Gemini-2.5 | Ovis2 | 2,537 | 2,220 | 4.61e-06* |
| Gemini-2.5 | Baichuan-Omni | 3,303 | 2,356 | 2.88e-36* |
| Qwen3-VL | Ovis2 | 2,159 | 2,573 | 1.93e-09* |
| Qwen3-VL | Baichuan-Omni | 2,948 | 2,732 | 0.0043* |
| Ovis2 | Baichuan-Omni | 2,501 | 1,871 | 1.86e-21* |

**Appendix Table 15.** Pairwise McNemar test results with Bonferroni correction for the Classification task.

| Model 1 | Model 2 | b | c | p-value |
|---|---|---|---|---|
| GPT-4o | Gemini-2.5 | 292 | 118 | 1.30e-17* |
| GPT-4o | Qwen3-VL | 225 | 65 | 9.93e-21* |
| GPT-4o | Ovis2 | 277 | 68 | 4.15e-29* |
| GPT-4o | Baichuan-Omni | 233 | 49 | 1.18e-27* |
| Gemini-2.5 | Qwen3-VL | 124 | 138 | 0.4219 |
| Gemini-2.5 | Ovis2 | 121 | 86 | 0.0181 |
| Gemini-2.5 | Baichuan-Omni | 216 | 206 | 0.6613 |
| Qwen3-VL | Ovis2 | 133 | 84 | 0.0011* |
| Qwen3-VL | Baichuan-Omni | 156 | 132 | 0.1753 |
| Ovis2 | Baichuan-Omni | 146 | 171 | 0.1777 |

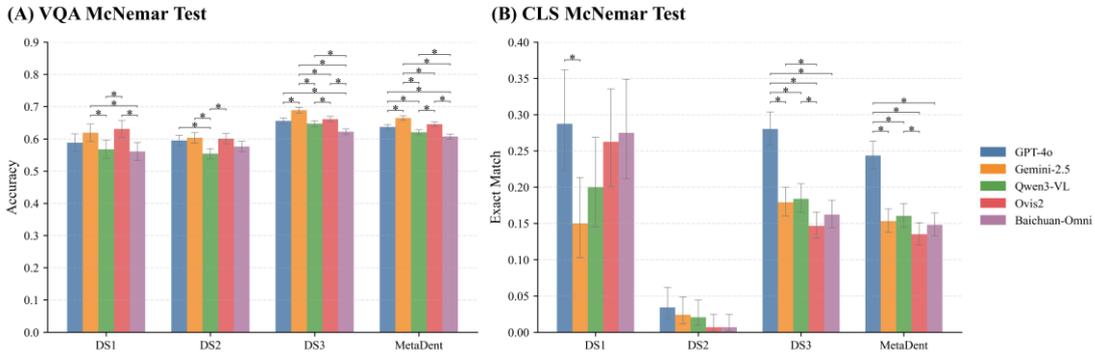

**Appendix Figure 6.** (A) VQA accuracy of the five VLMs evaluated on DS1, DS2, DS3, and the full MetaDent benchmark, together with pairwise McNemar tests; an asterisk (*) indicates a statistically significant difference between the corresponding model pair after Bonferroni correction. (B) Exact Match for the multi-label classification (CLS) task on DS1, DS2, DS3, and MetaDent, together with pairwise McNemar tests.